\documentclass[preprint,12pt,authoryear]{elsarticle}

\usepackage{amssymb}
\usepackage{amsmath}
\usepackage{subcaption}
\usepackage{makecell}
\usepackage{pifont}
\newcommand{\cmark}{\ding{51}}
\newcommand{\xmark}{\ding{55}}
\usepackage[colorlinks=true,linkcolor=blue,citecolor=blue,urlcolor=blue]{hyperref}
\usepackage[nameinlink,noabbrev]{cleveref}
\usepackage{tabularx}
\crefname{figure}{Fig.}{Figs.}   
\Crefname{figure}{Fig.}{Figs.}
\crefname{table}{Tab.}{Tabs.}

\journal{Engineering Applications of Artificial Intelligence}

\begin{document}

\begin{frontmatter}




\title{LSP-YOLO: A Lightweight Single-Stage Network for Sitting Posture Recognition on Embedded Devices}


\author[inst1]{Nanjun Li\fnref{eq1}\corref{cor1}} 
\ead{22225023@zju.edu.cn}
\author[inst1]{Ziyue Hao} 
\ead{haoziyue@zju.edu.cn}
\author[inst2]{Quanqiang Wang} 
\ead{Johnson@icheego.com}
\author[inst1]{Xuanyin Wang} 
\ead{xywang@zju.edu.cn}

\cortext[cor1]{Corresponding author. Email: 22225023@zju.edu.cn}
\affiliation[inst1]{organization={School of Mechanical Engineering, Zhejiang University},
            city={Hangzhou},
            postcode={310030}, 
            state={Zhejiang Province},
            country={China}}
\affiliation[inst2]{organization={Hangzhou Qige Zhilian Technology Co., Ltd.},%
            city={Hangzhou},
            postcode={310030}, 
            state={Zhejiang Province},
            country={China}}

\begin{abstract}
With the rise in sedentary behavior, health problems caused by poor sitting posture have drawn increasing attention.
Most existing methods, whether using invasive sensors or computer vision, rely on two-stage pipelines, which result in high intrusiveness, intensive computation, and poor real-time performance on embedded edge devices.
Inspired by YOLOv11-Pose, a lightweight single-stage network for sitting posture recognition on 
embedded edge devices termed LSP-YOLO was proposed.
By integrating partial convolution(PConv) and Similarity-Aware Activation Module(SimAM), a lightweight module, Light-C3k2, was designed to reduce computational cost while maintaining feature extraction capability.
In the recognition head, keypoints were directly mapped to posture classes through pointwise convolution, and intermediate supervision was employed to enable efficient fusion of pose estimation and classification.
Furthermore, 
a dataset containing 5,000 images across six posture categories was constructed for model training and testing.
The smallest trained model, LSP-YOLO-n, achieved 94.2\% accuracy and 251 Fps on personal computer(PC) with a model size of only 1.9 MB.
Meanwhile, real-time and high-accuracy inference under constrained computational resources was demonstrated on the SV830C + GC030A platform.
The proposed approach is characterized by high efficiency, lightweight design and deployability, making it suitable for smart classrooms, rehabilitation, and human–computer interaction applications.
\end{abstract}




\end{frontmatter}



\section{Introduction}
In modern society, increasing academic and occupational demands often force both adults and students to sit for long periods.
Prolonged improper sitting posture can result in various health problems, including scoliosis and lumbar spine injuries\citep{markova2024assessing,waongenngarm2015perceived,zemp2016occupational,szeto2002field}. Moreover, for students, such postural habits may further adversely affect their normal growth and development\citep{centemeri2024clinical,rosa20174}.
Therefore, real-time recognition of improper sitting posture and timely feedback for posture correction play a crucial role in preventing the aforementioned health problems. Moreover, sitting posture recognition has potential applications in fields such as autonomous driving and medical monitoring, enhancing the intelligence and human–computer interaction capabilities of related systems.

Currently, mainstream sitting posture recognition methods can be categorized into contact-based approaches using sensors\citep{tang2021upper,jiang2022knitted,tlili2022design,zhang2022privacy,tan2001sensing,huang2017smart,benocci2011context,wong2008trunk} and non-contact approaches based on computer vision\citep{vermander2024intelligent,hu2025research,liu2017healthy,kulikajevas2021detection,lin2023deep,min2018scene,chen2019sitting,liu20203d}. In most existing studies, both approaches typically adopt a two-stage recognition framework.
The core concept of this approach is to initially collect posture-related features during the perception stage. Sensor-based methods capture mechanical and positional features at the human–seat interface through pressure sensors, accelerometers, and other sensing devices.
In contrast, vision-based methods utilize human keypoint recognition models to extract keypoint features from images, including the eyes, nose, shoulders, and wrists.
In the subsequent recognition stage, the extracted features are input into an independent classifier to perform posture recognition and classification.
In theory, this two-stage recognition framework provides strong interpretability and high classification accuracy, demonstrating excellent performance, especially on high-performance PC platforms. However, the limitations of this approach are also considerable.
On the one hand, two-stage methods impose high computational demands, which hinder their direct deployment on embedded edge computing devices.
On the other hand, 
sensor-based contact approaches are highly intrusive, which reduces users' comfort.
For vision-based methods, deployment on edge devices typically relies on model compression techniques such as quantization and pruning. These processes introduce unavoidable quantization errors, reduce feature extraction precision, and ultimately impair posture classification performance.
In addition, in multi-person recognition scenarios such as smart classrooms and rehabilitation training, the computational cost and inference time of two-stage methods increase linearly with the number of detected subjects, further intensifying the computational burden.
These issues collectively hinder the widespread adoption and practical deployment of two-stage sitting posture recognition methods in real-world applications.

\begin{figure}[!h]
    \centering
    \captionsetup[sub]{font=small,labelformat=parens,labelsep=space}
    \begin{subfigure}{0.7\columnwidth}
        \includegraphics[width=\linewidth]{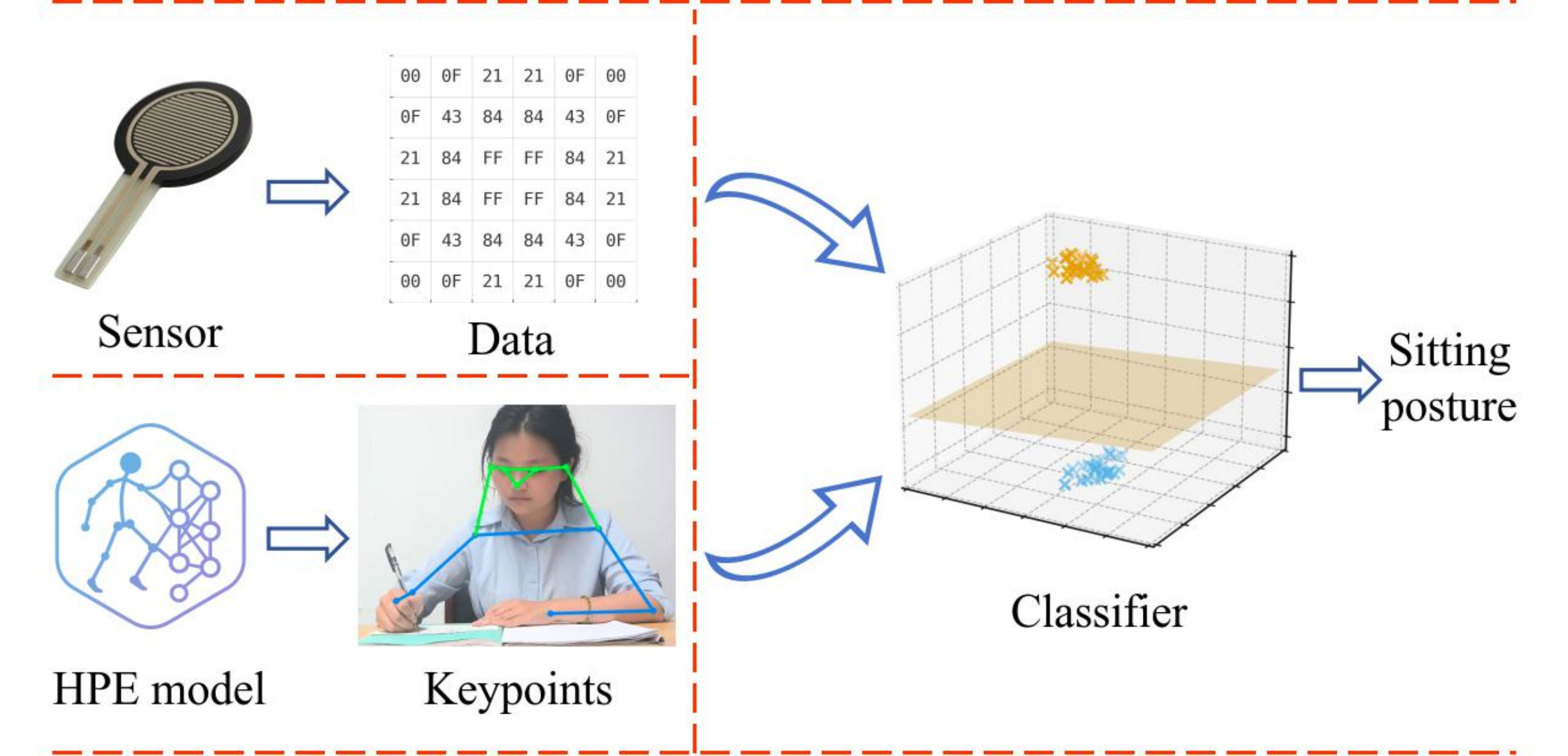}
        \caption{}
        \label{fig:comparison:a}
    \end{subfigure}
    \vspace{1.0em}
    \begin{subfigure}{0.7\columnwidth}
        \includegraphics[width=\linewidth]{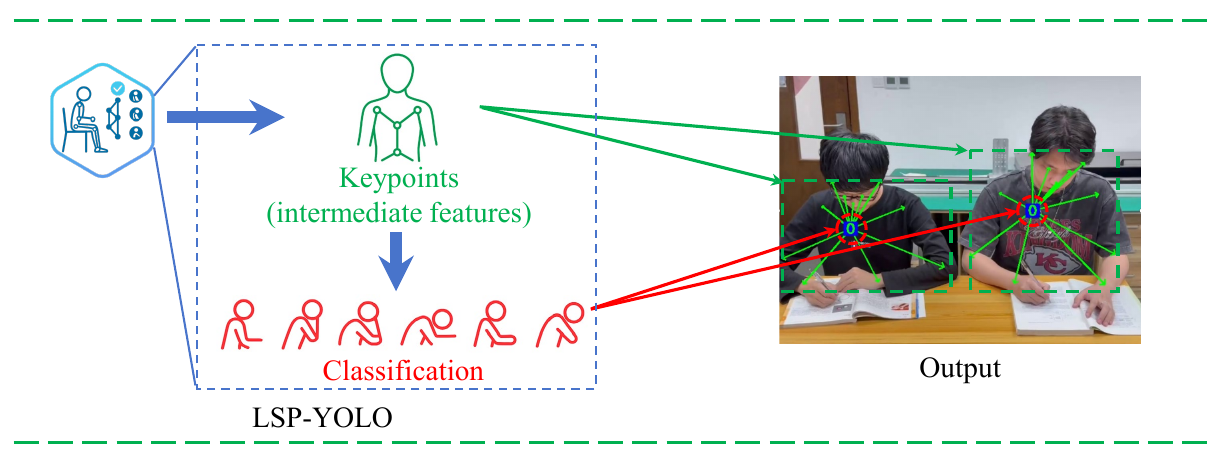}
        \caption{}
        \label{fig:comparison:b}
    \end{subfigure}
    \caption{Comparison between the conventional multi-stage pipeline and the proposed single-stage approach. 
  (a) Conventional two-stage method; (b) Proposed single-stage method.}
    \label{figure:comparison}
\end{figure}

To address these challenges and achieve high-precision, real-time recognition of poor sitting posture on embedded edge devices with constrained computational resources, 
this study proposed Lightweight Sitting Posture Recognition Network(LSP-YOLO) for edge devices with limited computing resources based on the YOLOv11-Pose\citep{maji2022yolo} model, 
aiming to perform efficient sitting posture recognition in an end-to-end, single-stage manner.
\Cref{figure:comparison} illustrates the differences between traditional methods and the proposed approach.

Our main contributions are as follows:
\begin{itemize}
    \item We proposed LSP-YOLO, a single-stage sitting posture recognition framework that integrates keypoint extraction and posture classification using pointwise convolution and intermediate supervision. This design reduced efficiency loss and error accumulation caused by feature storage, thereby significantly enhancing recognition performance.
    \item We designed Light-C3k2, a lightweight feature extraction module that combines partial convolution(PConv)\citep{chen2023run} with the parameter-free attention mechanism SimAM\citep{yang2021simam}, significantly reducing computational cost while maintaining efficient feature representation.
    \item We constructed a dataset of 5,000 images covering six types of poor sitting postures, and the proposed model achieved superior performance over existing methods in both speed and accuracy.
    \item We deployed and evaluated LSP-YOLO on the SV830C embedded edge platform, verifying its high-accuracy and low-latency performance in a resource-limited environment.
\end{itemize}

\section{Related work}
\subsection{Sensor-based sitting posture recognition methods}
Early sitting posture recognition methods primarily relied on sensor-based technologies for posture recognition.
These approaches collected pressure, inertial, or bioelectrical data and constructed corresponding models to achieve the recognition and classification of sitting posture.
For example, in 2017, Li et al. \cite{jian2017design} proposed a sitting posture pressure detection system based on a flexible tactile transducer using conductive rubber. The monitoring system employed a 64-element sensor array made of conductive rubber to collect and transmit sitting posture pressure signals. These signals were then processed by a host computer to generate a pressure distribution map of the human sitting posture.
In 2021, Farnan et al. Farnan\citep{farnan2021magnet} developed a low-cost wearable sitting posture recognition system with a unique structural design. The study introduced a magnet-integrated shirt that determines back posture using a magnetic sensor placed above the sternum, enabling the identification and continuous monitoring of the user’s sitting posture.
In 2022, Huang et al. \cite{huang2022transient} utilized flexible, degradable, transient, and highly sensitive MXene/cellulose nanofiber dual-type sensors—integrating pressure and bioelectric sensing—to collect pressure and electrocardiogram (ECG) signals.
A deep neural network was used as the classification model, enabling continuous and accurate detection of ECG, heart rate, sedentary behavior, and sitting posture.
In 2024, Yusoff et al. \cite{yusoff2024wheelchair} employed six pressure sensors to collect wheelchair users’ sitting posture data and applied three machine learning classification algorithms—support vector machine (SVM), random forests (RF), and decision tree (DT)—to classify the sitting posture. K-fold cross-validation was used for evaluation, and based on these methods, a sitting posture recognition system for wheelchair users was developed.

\subsection{Computer vision-based sitting posture recognition methods}
Several studies have utilized deep neural networks (DNNs) to directly extract features from images or video sequences for sitting posture classification.
For example, in 2021, Kulikajevas et al. \cite{kulikajevas2021detection} proposed a deep temporal hierarchical network built on the lightweight backbone MobileNetV2 \citep{howard2017mobilenets}. The model processed RGB-D frame sequences to generate semantic posture representations, achieving 91.47\% accuracy at 10 fps.
\cite{lin2023deep} developed an intelligent chair that supported the recognition of multiple sitting postures. Two depth cameras mounted on the chair collected the user’s full-body depth information, which was then processed by the lightweight EfficientNet \citep{koonce2021efficientnet} model to classify sitting postures.
Meanwhile, keypoint-based sitting posture classification methods, which rely on extracting human keypoint information from images, 
have also developed rapidly.
\cite{min2018scene} used a Kinect camera to extract human skeletal keypoints and fused them with background information detected by Faster R-CNN \citep{ren2016faster}.
A Gaussian mixture behavior clustering method was then applied to process the data, providing rich semantic information and enabling high-accuracy sitting posture recognition in screen-reading scenarios.
Chen et al. \citep{chen2019sitting} proposed a classroom student posture recognition system based on OpenPose\citep{cao2019openpose}.
The system employed surveillance cameras to detect student postures, classified sitting postures using Convolutional Neural Network(CNN), and achieved over 90\% recognition accuracy.
\cite{liu20203d}proposed a skeleton-based posture detection model named 3D PostureNet, which applied skeleton data normalization and Gaussian voxelization to enhance the robustness of posture classification.

\section{Methods}
\subsection{YOLOv11-Pose}
\begin{figure}[!h]
    \centering
    \includegraphics[width=0.9\textwidth]{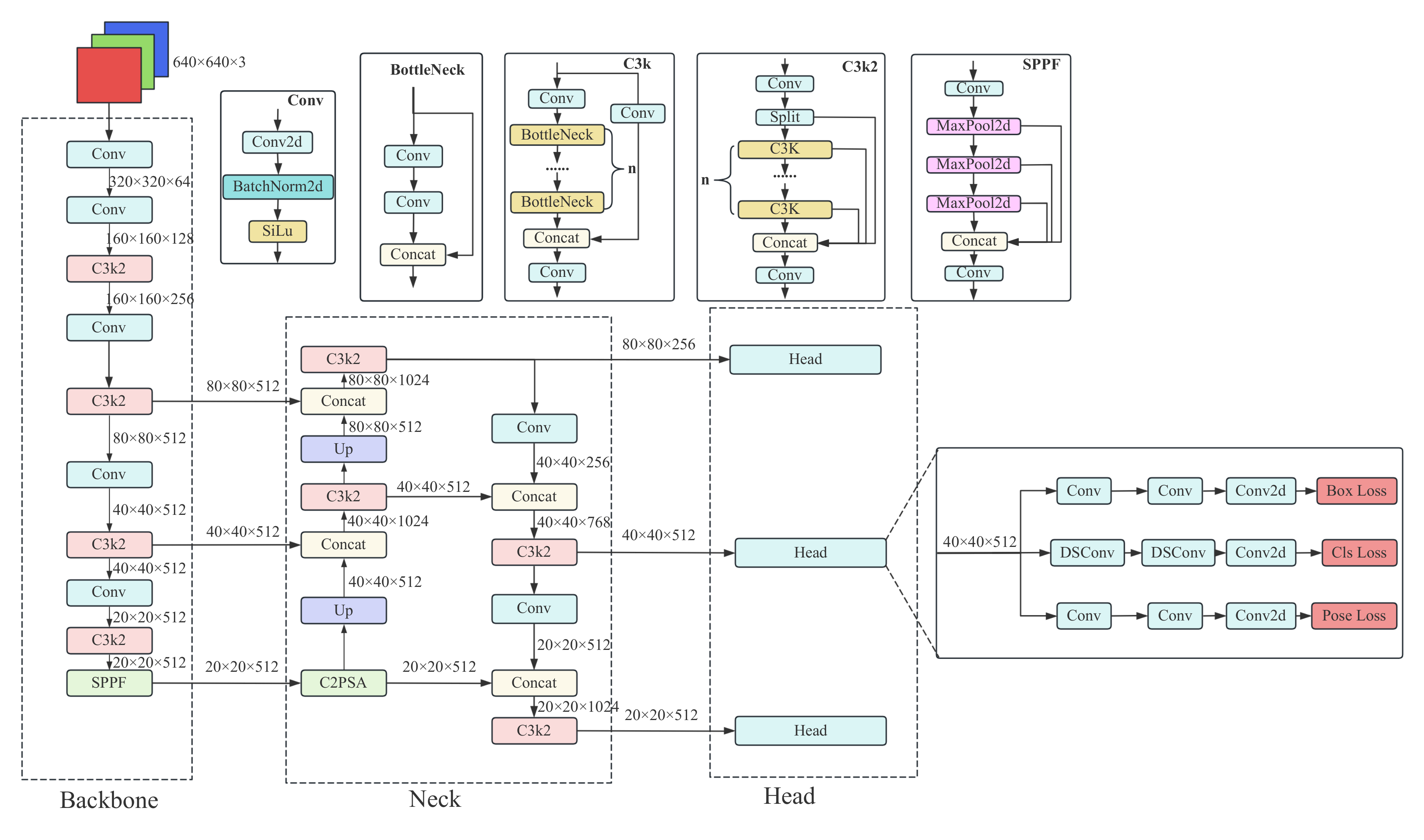}
    \caption{The network structure of YOLOv11-Pose}
    \label{figure:YOLOv11-Pose}
\end{figure}

YOLOv11-Pose is a lightweight human pose estimation network built on the YOLOv11 architecture.
The model inherits YOLOv11’s efficient feature extraction backbone and multi-scale detection head, and introduces structural modifications to better address the keypoints evaluation task, thereby enabling efficient end-to-end pose estimation.
The network architecture is illustrated in \cref{figure:YOLOv11-Pose}.

Its backbone utilizes the C3k2 module and SPPF block for feature extraction.
The features extracted from three key layers are fed into the neck for multi-scale feature fusion, fully leveraging shallow spatial and deep semantic information to achieve complementary feature enhancement.
The fused features are further fed into the head to directly regress target classes, bounding boxes, and keypoint coordinates, enabling integrated prediction of pose estimation and object detection.
In addition, YOLOv11-Pose continues the training and inference strategy of YOLOv10. By directly regressing the human keypoint coordinate vectors, it avoids dependence on post-processing operations such as Non-Maximum Suppression(NMS), achieving an end-to-end differentiable training process.
This design not only improves inference efficiency but also reduces deployment complexity, making it well suited for real-time applications on resource-constrained edge devices.
The differentiability of the network enables the deep integration of human pose estimation with downstream sitting posture classification, and lays the foundation for the keypoint-based end-to-end sitting posture recognition method proposed in this study.

\subsection{Single-stage sitting posture recognition network: LSP-YOLO}
Based on the YOLOv11-Pose architecture, this study proposed a single-stage real-time sitting posture recognition network targeting embedded edge devices — Lightweight Sitting Posture YOLO (LSP-YOLO).

\begin{figure}[!h]
    \centering
    \includegraphics[width=\textwidth]{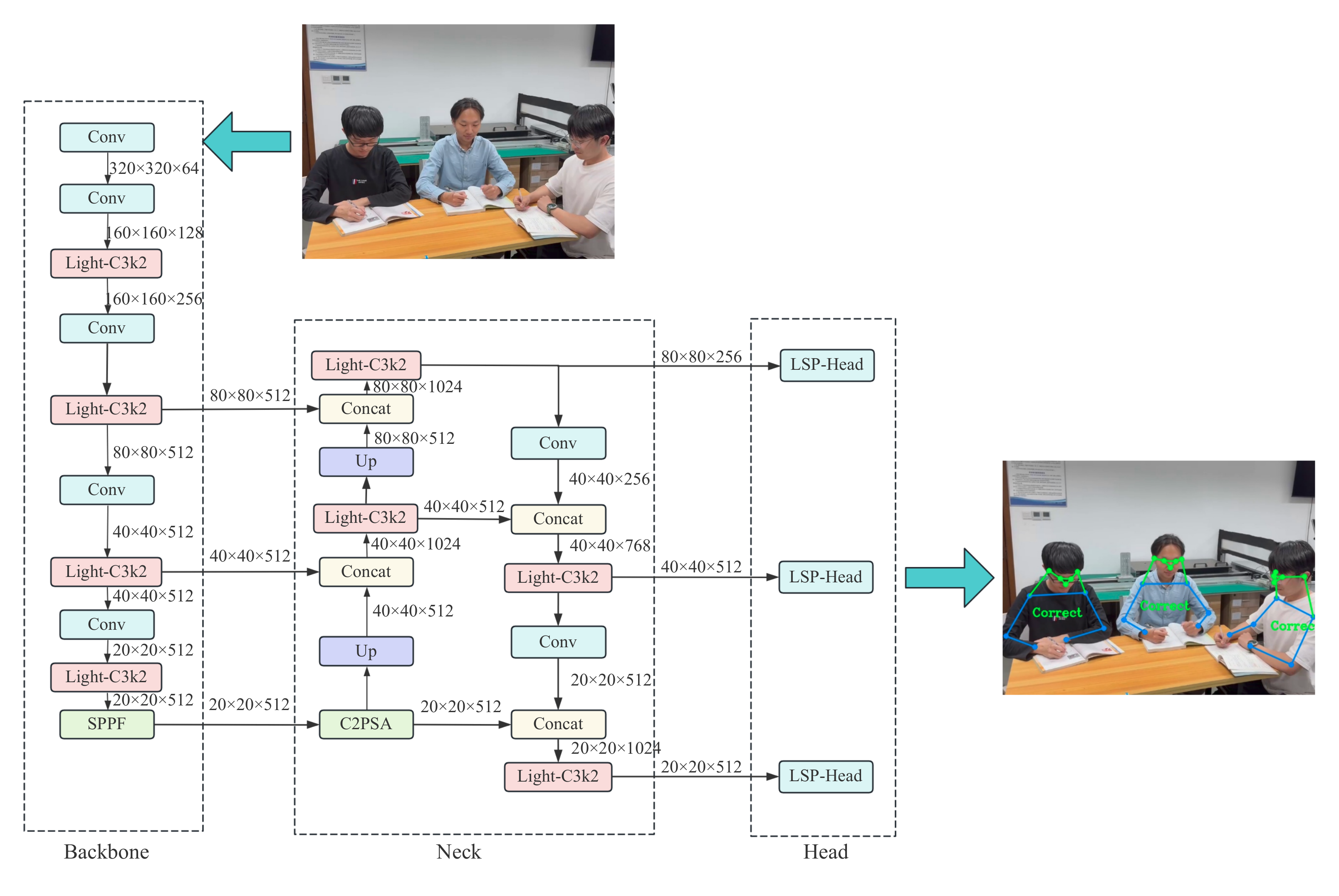}
    \caption{The overall structure of LSP-YOLO}
    \label{figure:LSP-YOLOv11}
\end{figure}

Specifically, we introduced Point Convolution to map the keypoint vectors extracted by the network to sitting posture class probability distributions, thereby seamlessly integrating the sitting posture classification task into the keypoints evaluation network with minimal computational overhead.
At the same time, to ensure that the classification module received high-quality keypoints information, we introduced intermediate supervision into the keypoints evaluation branch.
By precisely supervising intermediate features, we effectively enhanced the coupling between pose and classification features, enabling the sitting posture recognition process to achieve genuine end-to-end training and inference.
At the same time, to ensure that the classification module received high-quality keypoint information, we introduced intermediate supervision to the keypoints evaluation branch. By applying precise supervision to intermediate features, we effectively enhanced the information coupling between pose features and classification features, enabling the entire sitting posture recognition process to be trained and inferred in an end-to-end manner.
In addition, to further improve the inference efficiency of the network on embedded devices, we introduced Partial Convolution into the baseline backbone structure and integrated the SimAM energy function, thereby proposing a lightweight feature extraction module, Light-C3k2. This module reduced the computational load by suppressing redundant feature maps and weighting important features, which significantly lowered the model’s computational cost while maintaining detection accuracy and effectively improved the network’s performance on embedded edge devices.
The overall architecture of LSP-YOLO is shown in  \cref{figure:LSP-YOLOv11}.

\subsection{Sitting posture classification with Point Convolution and LSP-Head}
Previous studies usually employed separate classifiers (such as support vector machines or random forests) to classify predicted keypoint coordinates.
This approach decoupled the detection and classification processes, 
which tended to cause error accumulation during feature transmission and resulted in low inference efficiency, thereby making it difficult to meet the dual requirements of limited computational resources and real-time performance on embedded devices.

We noticed that in the YOLOv11-Pose inference 
pipeline, the head part outputs grid-like feature
$F\in \mathbb{R}^{C\times H\times W}$.
which contains the confidence scores of 
${H\times W}$ candidate predictions and their associated keypoint vectors. These vectors are not affected by background information and able to accurately reflect 
the human pose characteristics.
In addition, each vector set is independent and non-interfering, thus well suited as input for posture classification.

\begin{figure}[!h]
    \centering
    \includegraphics[width=0.7\textwidth]{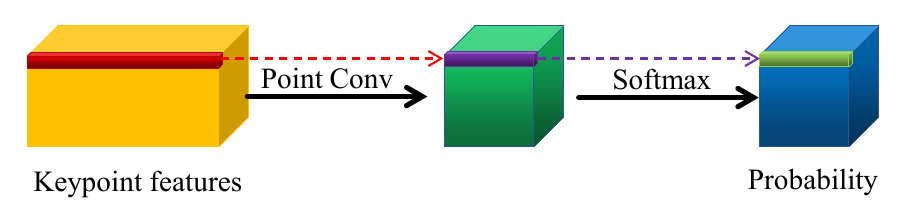}
    \caption{Point convolution–based sitting posture classification}
    \label{figure:Point Conv}
\end{figure}

\begin{figure}[!h]
    \centering
    \captionsetup[sub]{font=small,labelformat=parens,labelsep=space}
    \begin{subfigure}[c]{0.48\columnwidth}
        \centering
        \includegraphics[width=\linewidth]{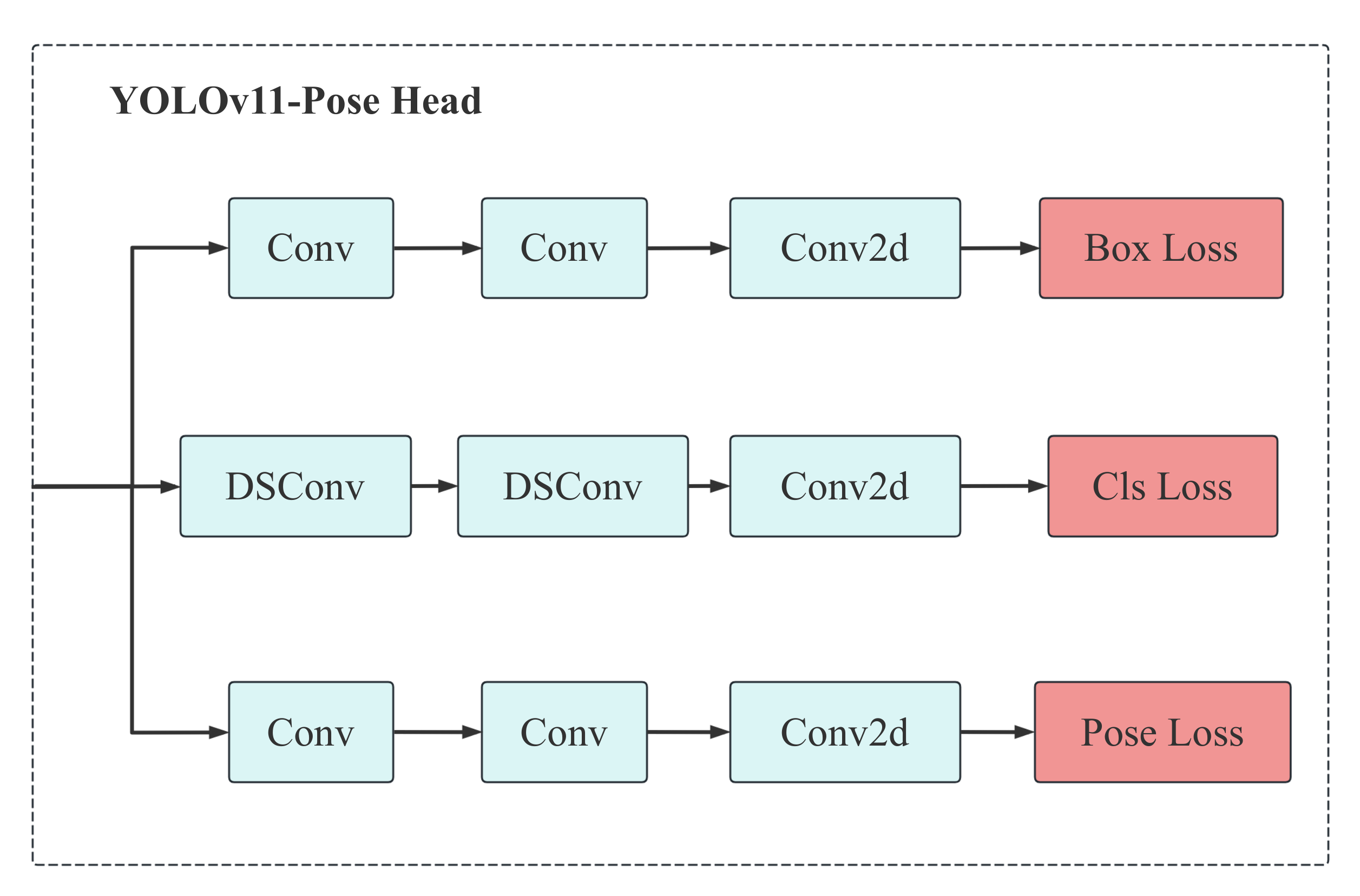}
        \caption{}
        \label{fig:head comparison:a}
    \end{subfigure}
    \vspace{1.0em}
    \begin{subfigure}[c]{0.48\columnwidth}
        \centering
        \includegraphics[width=\linewidth]{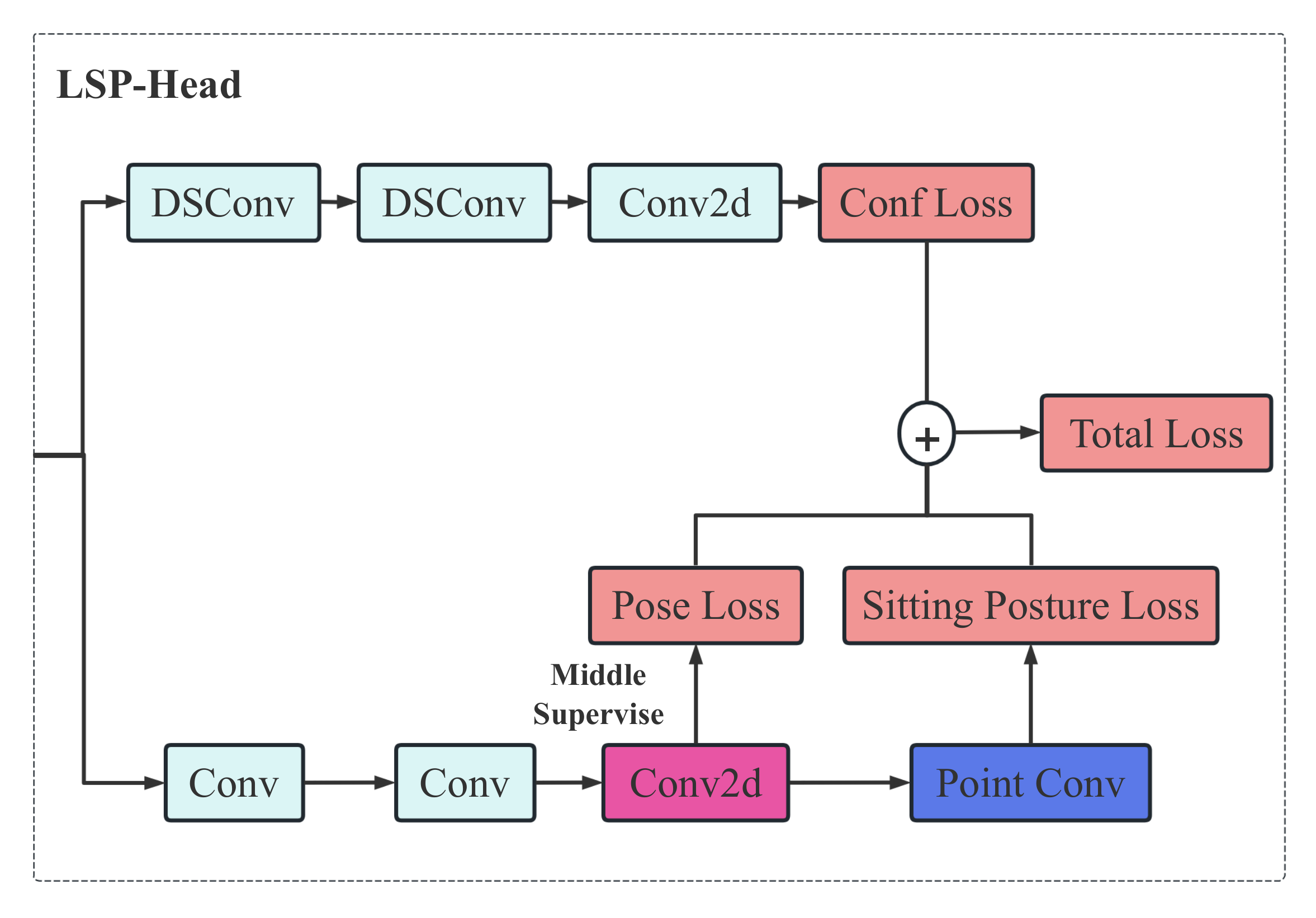}
        \caption{}
        \label{fig:head comparison:b}
    \end{subfigure}
    \caption{Comparison between YOLOv11-Pose Head and the proposed LSP-Head. (a) YOLOv11-Pose Head; (b) Proposed LSP-Head.}
    \label{figure:comparison 1}
\end{figure}

\begin{figure}[!h]
    \centering
    \includegraphics[width=0.9\textwidth]{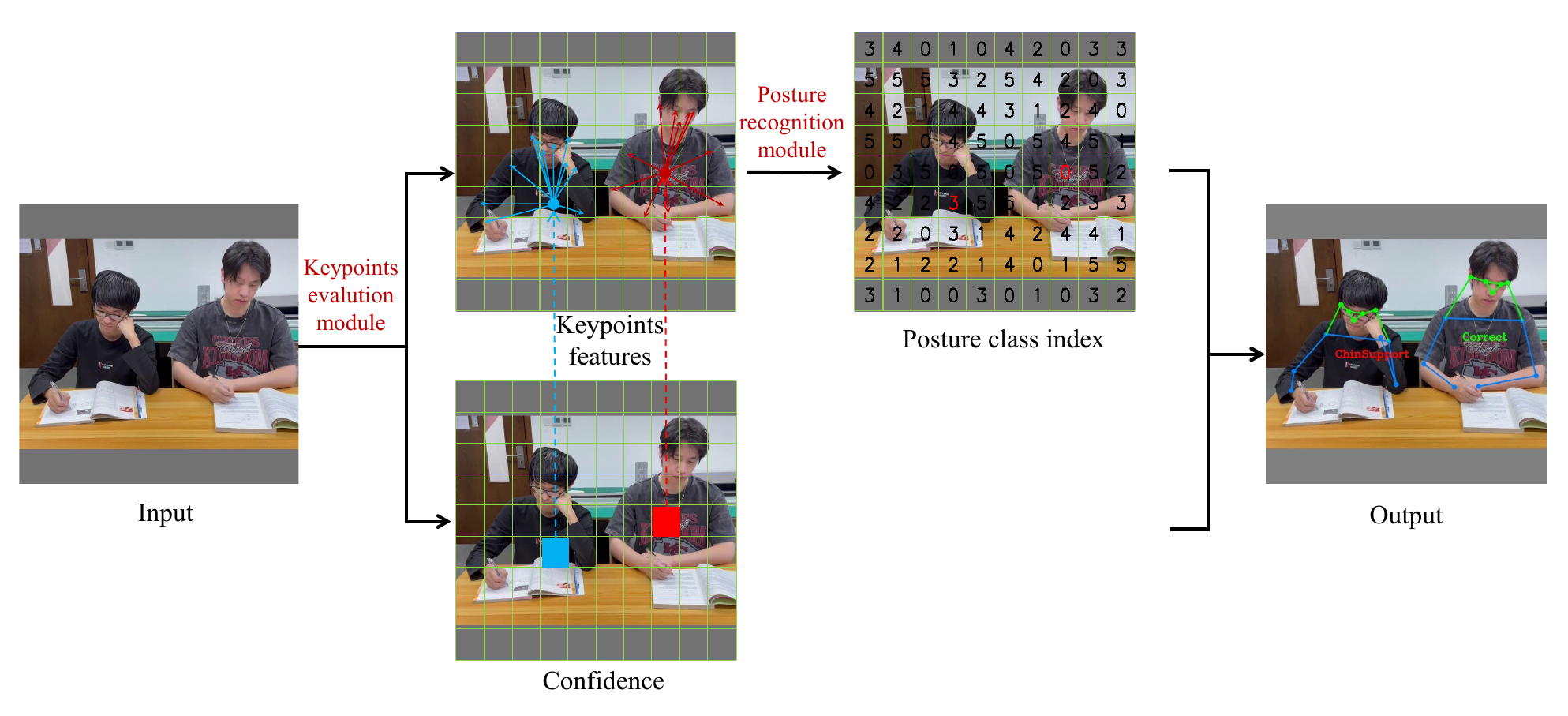}
    \caption{Overall inference pipeline of the proposed LSP-YOLO for sitting posture recognition}
    \label{figure:Inference}
\end{figure}

Based on this, we designed an efficient sitting posture classification module using point convolution (1×1 convolution).
Specifically, the keypoint vector $\hat{K}$ at each grid was linearly mapped with a 1×1 convolution to obtain category scores $S$:
\begin{equation}
    S = \mathrm{Conv}_{1 \times 1}(\hat{K})
\end{equation}
Then softmax function was applied to normalize these scores and obtain the posture category distribution:
\begin{equation}
    \hat{p_i} = \frac{e^{s_i}}{\sum_{i=1}^{6} e^{s_i}}
\end{equation}
Where $\hat{p_i}$ denotes the probability of class $i$. 
$s_i$ is the corresponding convolution output.
The process is shown in \cref{figure:Point Conv}.
Under this design paradigm, the human posture vector was repurposed as an intermediate representation,
with the final output being the sitting posture classification result.
With this module added, the head structure comparison between YOLOv11-Pose and LSP-YOLO is shown in \cref{figure:comparison 1}.
The overall inference pipeline of the proposed LSP-YOLO is shown in \cref{figure:Inference}. The input image is first processed by the keypoints evaluation module to extract upper-body keypoints, followed by the sitting posture classification Module that maps keypoint features to posture categories through pointwise convolution. The final output visualizes detected keypoints and corresponding posture labels.
\subsection{Light-C3k2}
\begin{figure}[!h]
    \centering
    \includegraphics[width=0.9\textwidth]{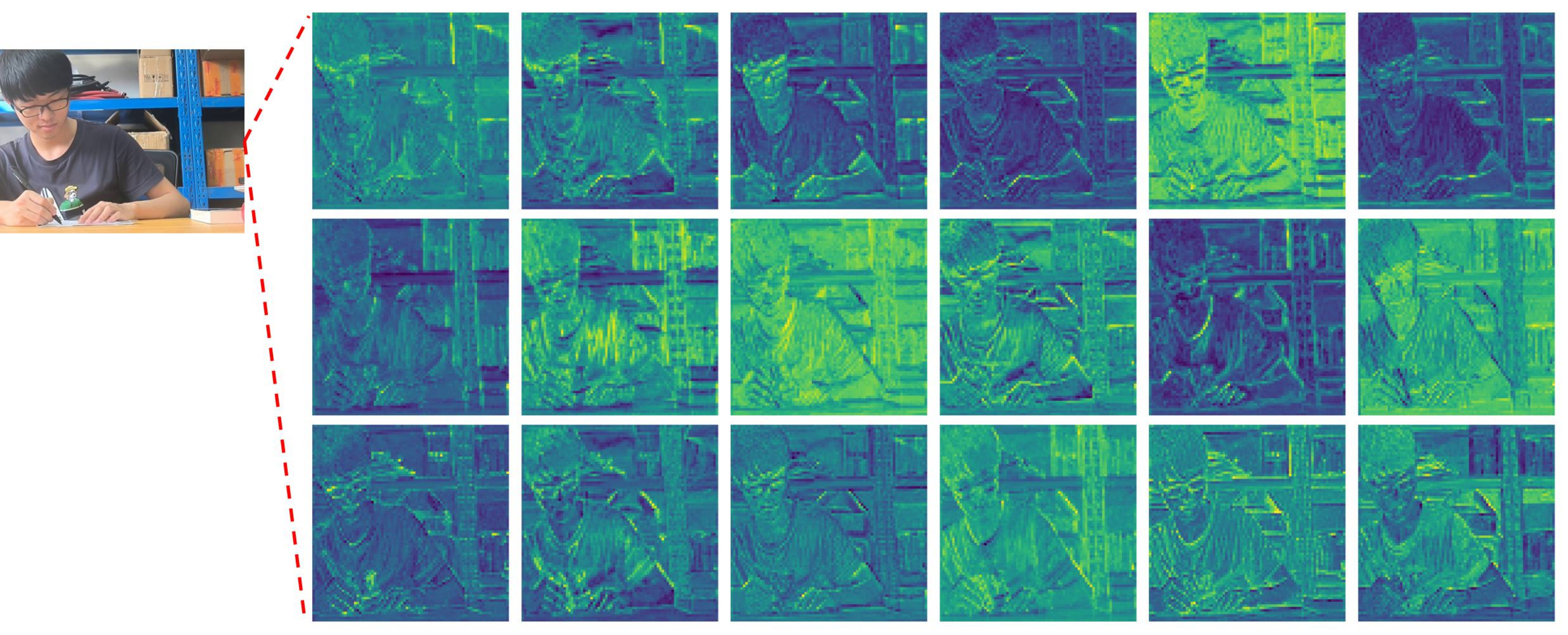}
    \caption{The feature maps exhibiting similarity}
    \label{figure:repeated_features}
\end{figure}

In conventional CNNs, all input pixels participate in convolution, often producing many redundant and highly similar feature maps.
\cref{figure:repeated_features} illustrates the similarity among feature maps generated within the same convolutional layer.

Such redundancy has little impact on high-performance devices and may aid finer-grained information extraction.
However, on resource-constrained devices, these operations would cause high compute and power overhead, with poor accuracy–cost trade-offs.

\begin{figure}[!h]
    \centering
    \includegraphics[width=0.9\textwidth]{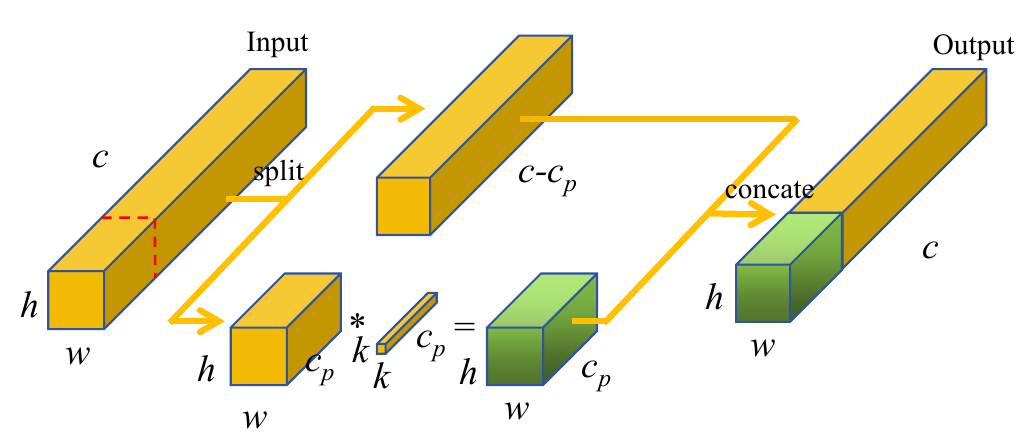}
    \caption{The computational process of PConv}
    \label{figure:PConv}
\end{figure}

To address this issue, we introduced the concept of partial convolution (PConv), in which only a subset of feature channels is convolved while the remaining channels remain unchanged. The computational process is illustrated in \cref{figure:PConv}.

Specifically, the number of parameters to perform partial convolution is:
\begin{equation}
    (r \cdot c)^2 \times k^2
\end{equation}

The computational cost for a single partial convolution operation is:
\begin{equation}
    h \times w \times (r \cdot c)^2 \times k^2
\end{equation}

\begin{figure}[!h]
    \centering
    \includegraphics[width=0.5\textwidth]{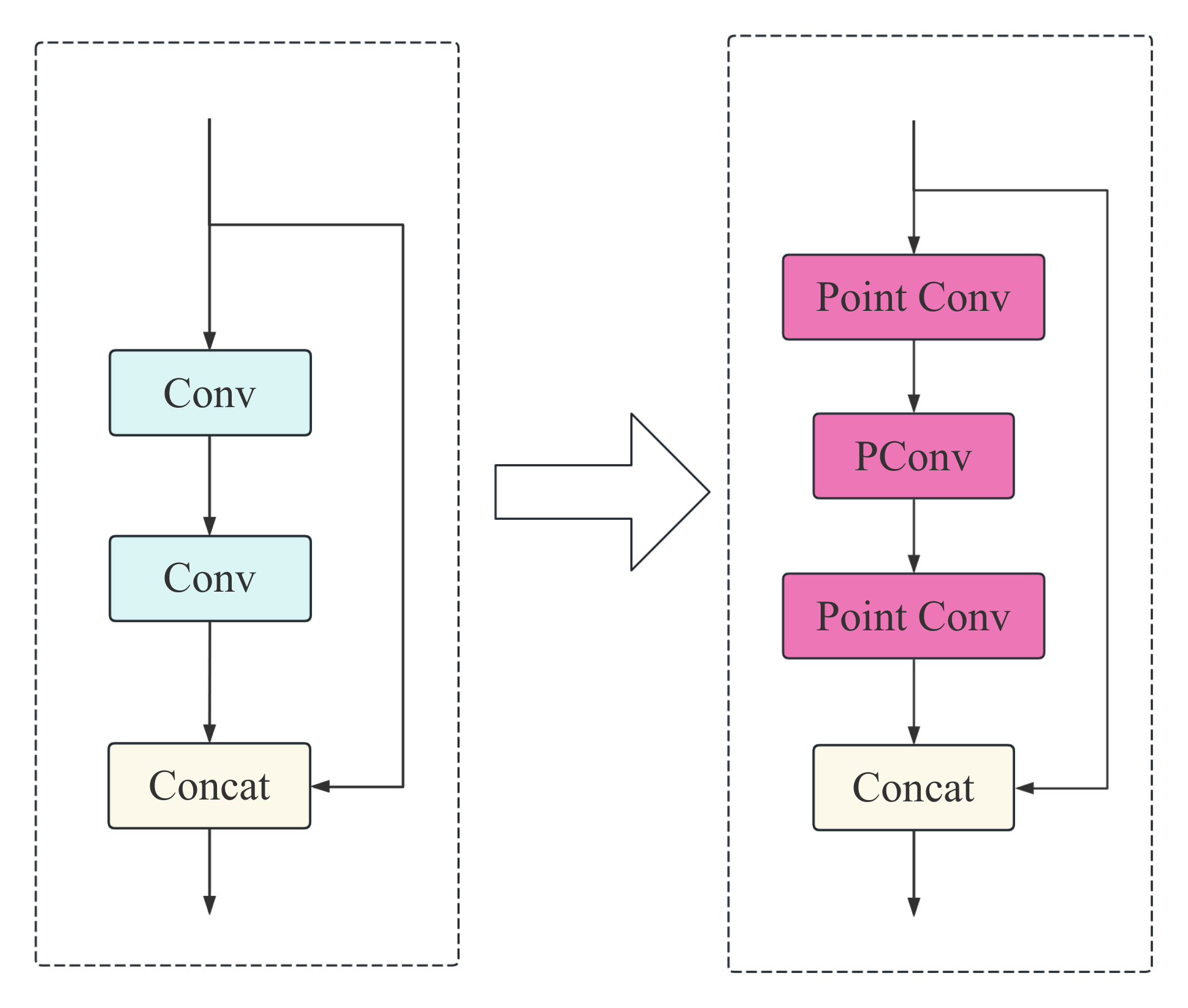}
    \caption{Optimization of the BottleNeck structure}
    \label{figure:Light-Block}
\end{figure}

The parameter size and computational cost of standard convolution are respectively:
\begin{equation}
    c \times k^2
\end{equation}

\begin{equation}
    h \times w \times c^2 \times k^2
\end{equation}
Where $h$ and $w$ denote the height and width of the feature map, $c$ represents the number of feature channels, $k$ is the kernel size, and $r$ denotes the ratio of channels participating in partial convolution.
Compared with standard convolution, partial convolution reduces storage and memory usage by a factor of $(1 - r^2)$.
This concept was incorporated into the BottleNeck block, where the first standard convolution was replaced by PConv. Feature fusion and extraction were achieved through two consecutive pointwise convolutions, effectively reducing computational cost while preserving the quality of the output features.
The structure is illustrated in \cref{figure:Light-Block}.

\begin{figure}[!h]
    \centering
    \includegraphics[width=0.7\textwidth]{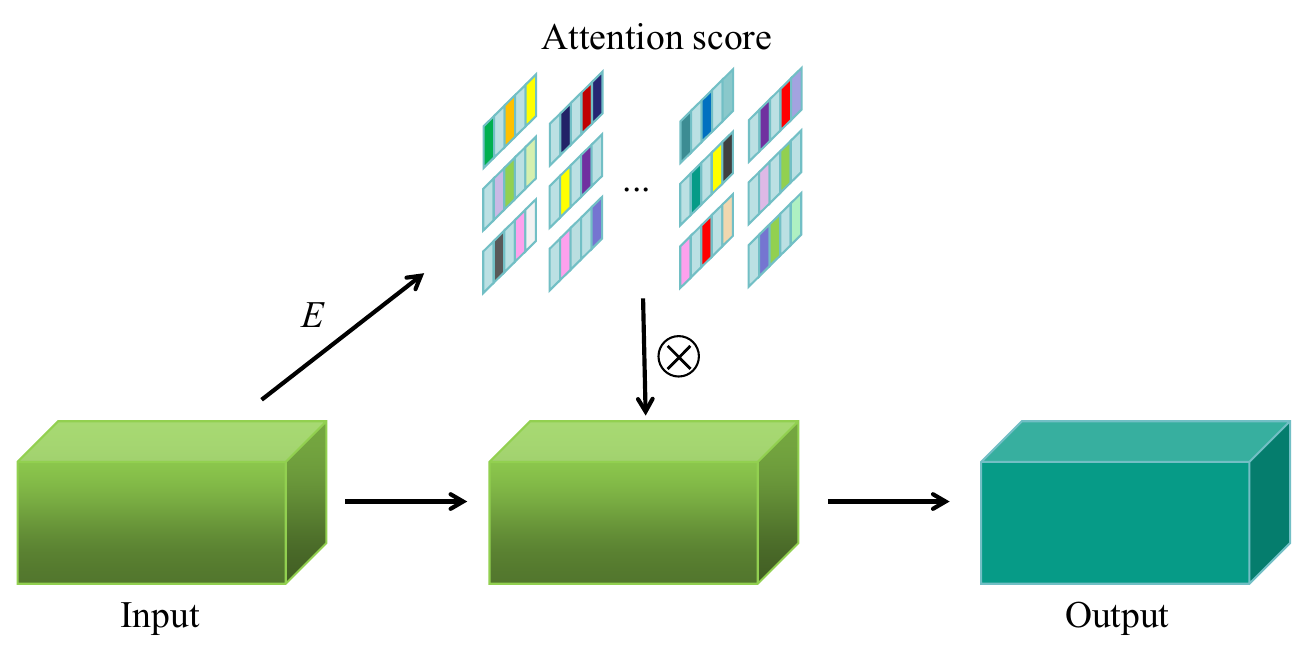}
    \caption{The computational process of SimAM}
    \label{figure:SimAM}
\end{figure}

Since PConv only operated on part of the input features, its feature utilization was reduced compared with standard convolution. To address this, a parameter-free and computationally simple SimAM attention module was introduced to emphasize important features and enhance the model’s discriminative ability.
However, conventional attention mechanisms, such as SE\citep{hu2018squeeze} and CBAM\citep{woo2018cbam}, usually introduce additional parameters.
Considering the limited computational resources of edge devices,
we adopt the SimAM attention module, which is parameter-free and computationally efficient.
SimAM evaluates neuron importance by optimizing an energy function and produces 3D attention weights from the input features.
The energy function is defined as follows:
\begin{equation}
    E(w_t, b_t, y, x_i) = \frac{1}{M-1} \sum_{i=1}^{M-1} \left[-1 - (w_t x_i + b_t)\right]^2 + \left[1 - (w_t t + b_t)\right]^2 + \lambda w_t^2
\end{equation}
Where $x_i$ denotes neurons in the same channel except the target neuron $t$; $w_t$ and $b_t$ are linear transformation parameters; $\lambda w_t^2$ is the regularization term to prevent overfitting; and $M$ is the number of neurons in the channel.
The energy function aimed to distinguish the target neuron from background neurons by driving the target activation toward $1$ and others toward $-1$.
A smaller energy value indicates a more important target neuron. By solving the closed-form of $w_t$ and $b_t$ and substituting them into the equation, the target neuron’s energy value can be obtained:
\begin{equation}
    e_t^* = \frac{4(\hat{\sigma}^2 + \lambda)}{(t - \hat{\mu})^2 + 2\hat{\sigma}^2 + 2\lambda}
\end{equation}
Then the attention score for each neuron was defined as:
\begin{equation}
    a = \mathrm{sigmoid}\left(\frac{1}{e_t^*}\right)
\end{equation}
Finally, for the input $X \in \mathbb{R}^{C \times H \times W}$,
after applying SimAM, 
the output is:
\begin{equation}
    \tilde{X}=A\cdot X
\end{equation}
where $A \in \mathbb{R}^{C \times H \times W}$ represents the attention scores $a$ derived from $X$.
\cref{figure:SimAM} illustrates the computational process of SimAM.

\begin{figure}[!h]
    \centering
    \includegraphics[width=0.8\textwidth]{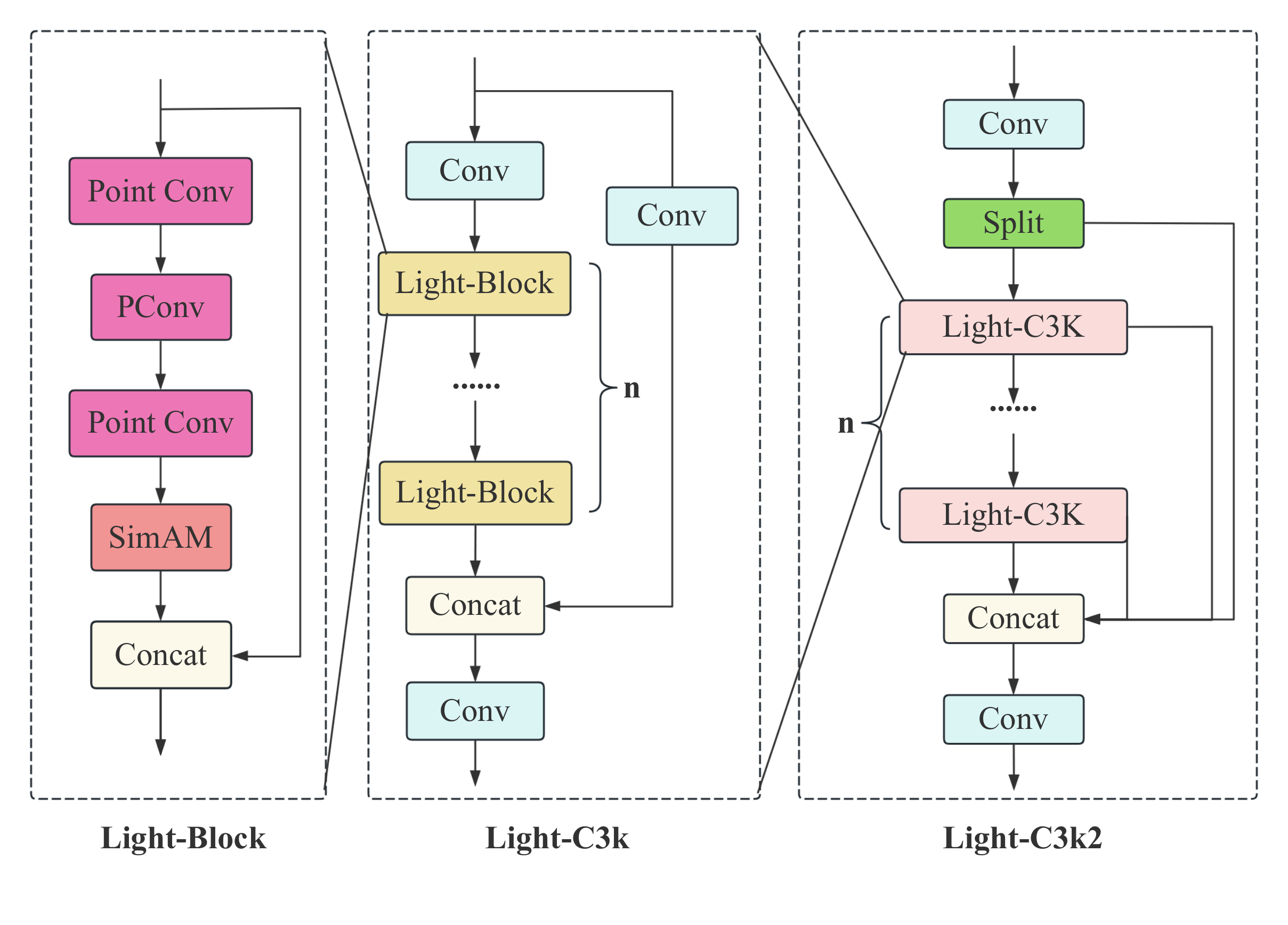}
    \caption{The structure of Light-C3k2}
    \label{figure:Light-C3k2}
\end{figure}

SimAM attention was applied to each BottleNeck containing PConv to reweight key neuron information and enhance the module’s feature extraction capability without introducing additional parameters.
The optimized units and modules were named Light-BottleNeck, Light-C3k, and Light-C3k2.
Their structures are illustrated in \cref{figure:Light-C3k2}.

\subsection{Design of the loss function}
Since each grid cell feature at the head output represents a predicted sample, it is necessary to first determine whether the grid corresponds to a positive sample by evaluating its confidence. The binary cross-entropy (BCE) loss function was adopted as the confidence loss function, defined as:
\begin{equation}
    L_{conf}=\sum_{n=1}^{N}BCE(p_{conf},t_{conf})
\end{equation}
Where $N$ denotes the number of grids, $p_{\text{conf}}$ represents the confidence predicted for a single grid, and $t_{\text{conf}}$ indicates whether the grid corresponds to a positive sample, taking the value of 1 if positive and 0 otherwise.

In LSP-YOLO, human keypoint information serves as intermediate features and is fed into point convolution for posture classification.
Thus, keypoint accuracy is critical to the final classification result. To ensure accurate keypoint prediction, the Object Keypoint Similarity (OKS) loss was introduced as intermediate supervision, defined as follows:
\begin{equation}
    L_{\mathrm{oks}}=1-\frac{\sum_{i}exp{({-d_i^2/2s^2k_i^2})}\delta(v_i>0)}{\sum_{i}\delta(v_i>0)}
\end{equation}
where $d_i$ denotes the Euclidean distance between the predicted and ground-truth positions of the $i$-th keypoint; $s$ is the scale factor, defined as the square root of the bounding box area; $k_i$ is the importance weight derived from the standard deviation of ground-truth annotations to reflect keypoint importance and scale uncertainty; and $v_i$ is the visibility label of the $i$-th keypoint.

For the final posture classification task, the binary cross-entropy (BCE) loss was used as the classification loss, defined as:
\begin{equation}
    L_{cls}=\sum_{n=1}^{N}BCE(p_{cls},t_{cls})
\end{equation}
where $p_{cls}$ represents the predicted class probability at a single grid; and $t_{cls}$ denotes the ground-truth class label.

In summary, the total loss was defined as the weighted sum of the three loss components:
\begin{equation}
    L_{sum}=\alpha L_{conf} + \beta L_{OKS} + \gamma L_{cls}
\end{equation}
where $\alpha$, $\beta$, and $\gamma$ denote the weights of the confidence loss, keypoint loss, and posture classification loss, respectively, and were set to 1, 12, and 4 in this study.

\section{Experiments and Results}
\subsection{Settings}
\subsubsection{Dataset preparation}
\begin{figure}[!h]
    \centering
    \begin{subfigure}{0.3\columnwidth}
        \includegraphics[width=\linewidth]{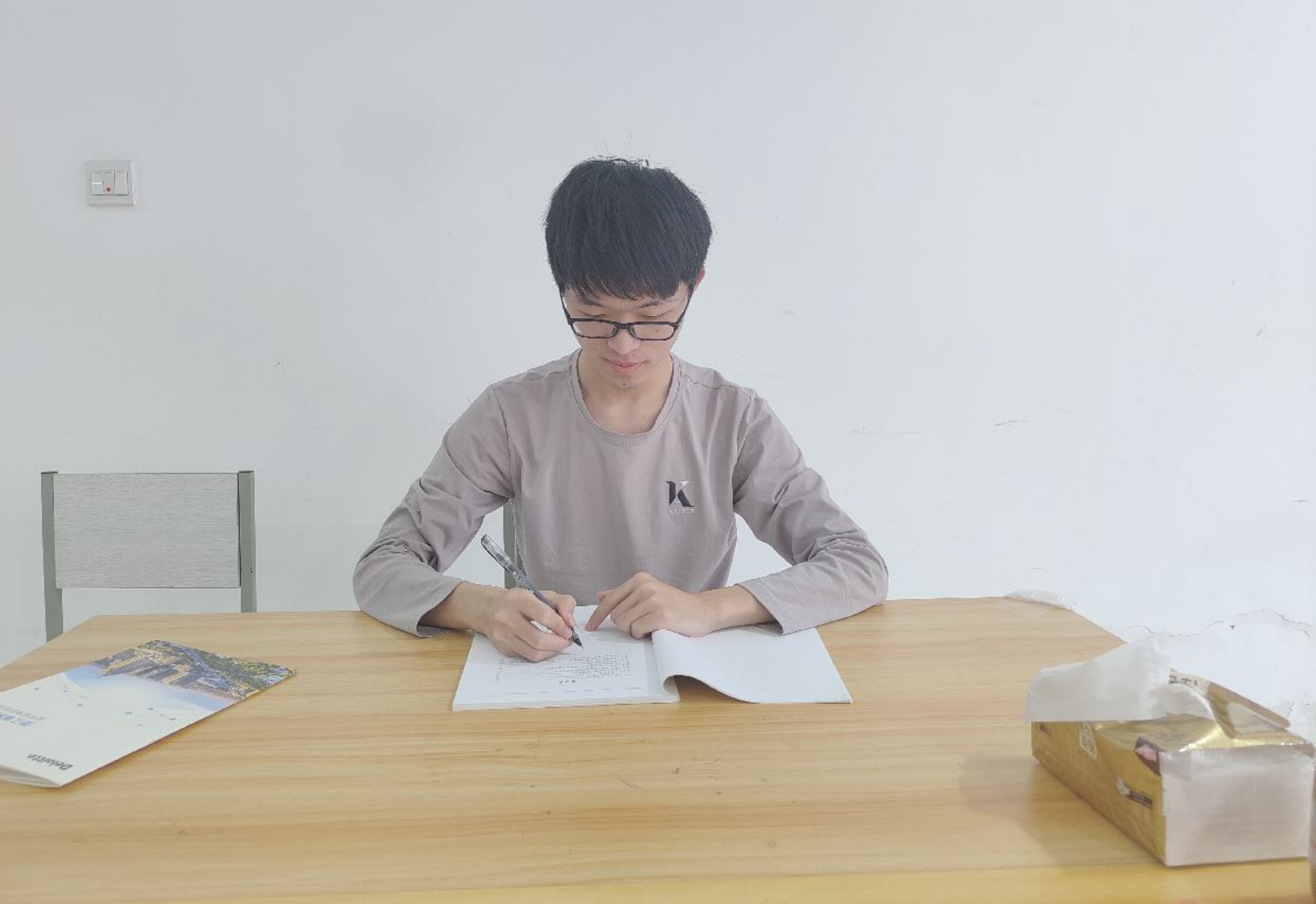}
        \caption{Correct}
    \end{subfigure}
    \begin{subfigure}{0.3\columnwidth}
        \includegraphics[width=\linewidth]{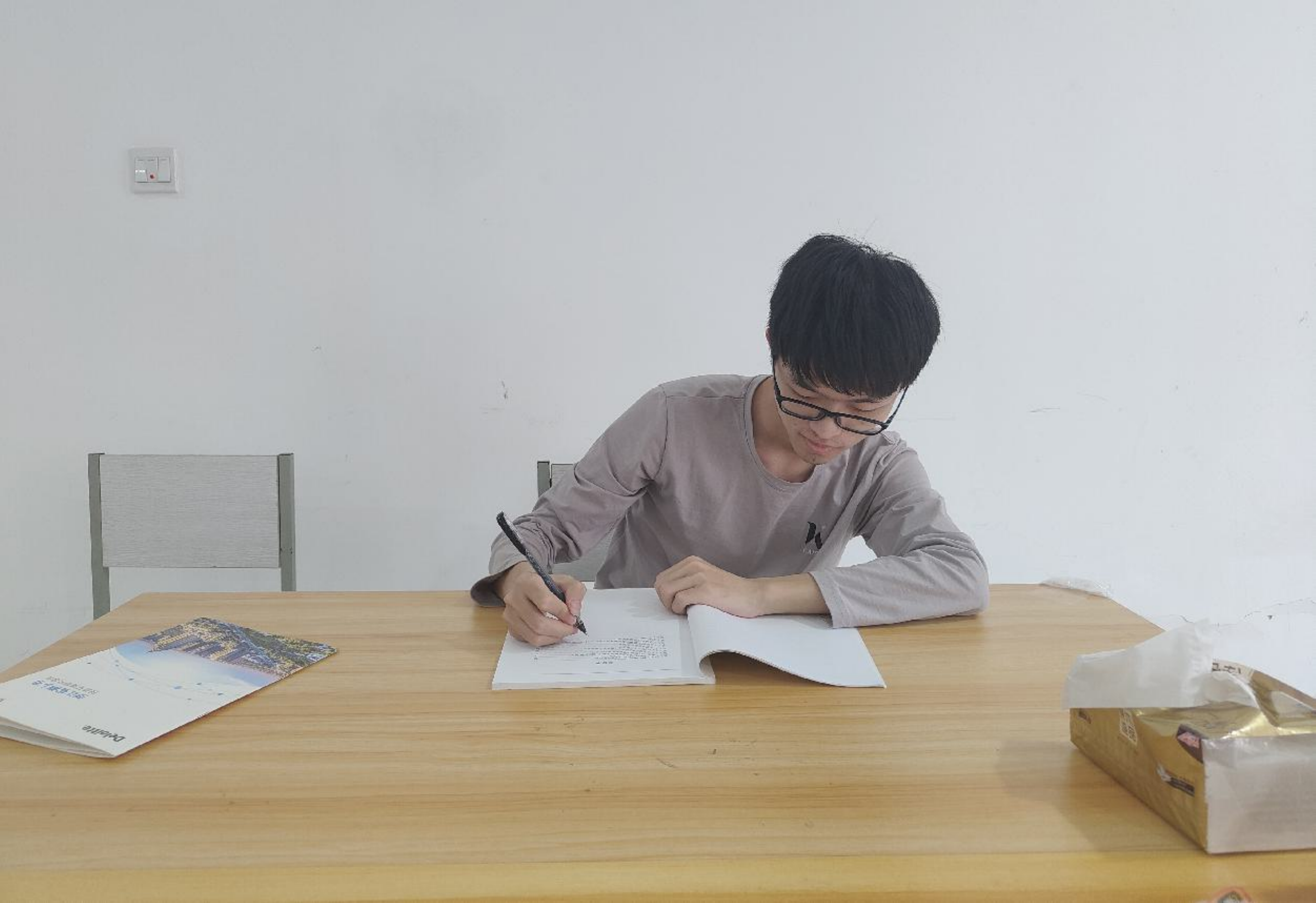}
        \caption{LeanLeft}
    \end{subfigure}
    \begin{subfigure}{0.3\columnwidth}
        \includegraphics[width=\linewidth]{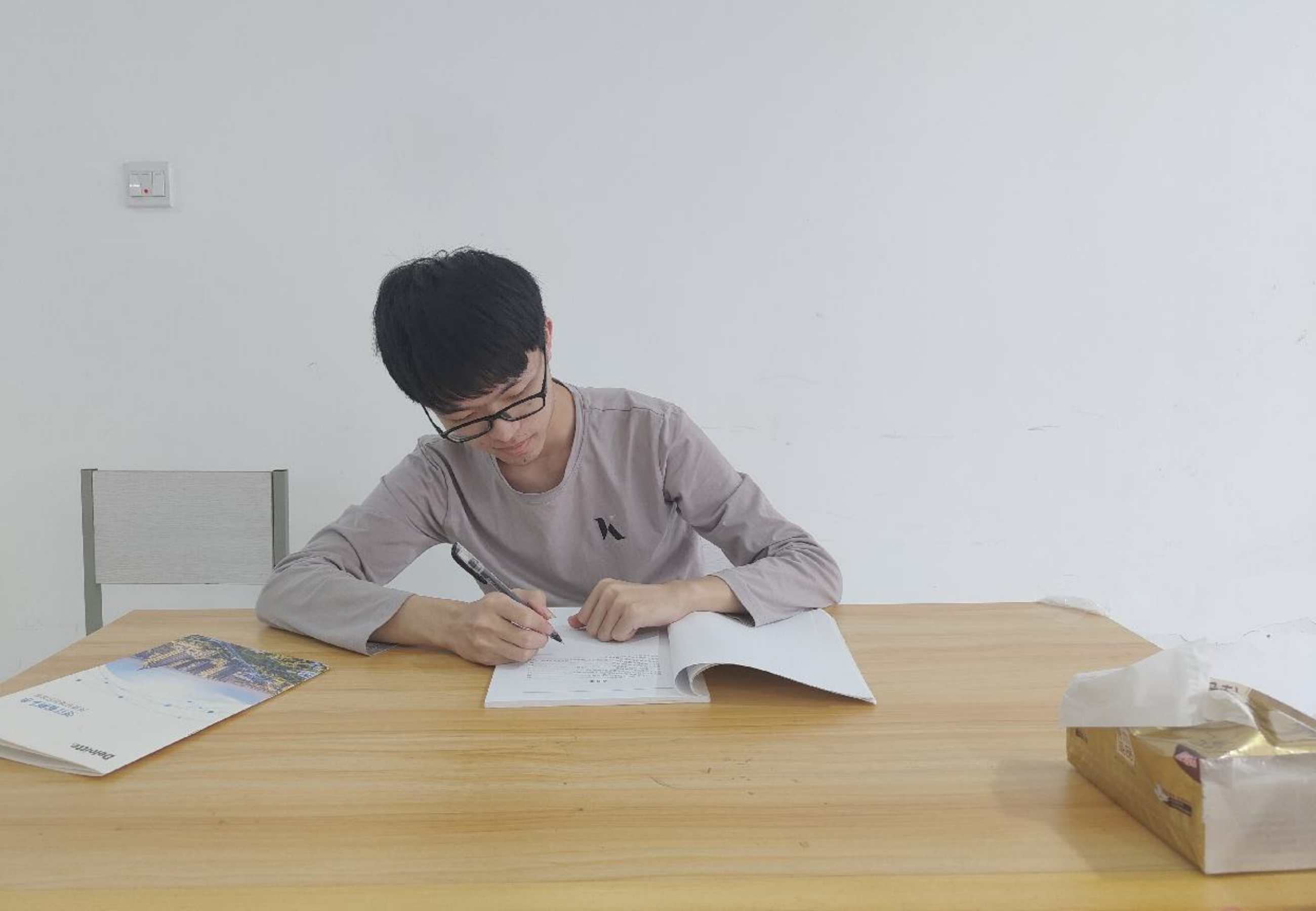}
        \caption{LeanRight}
    \end{subfigure}
    \begin{subfigure}{0.3\columnwidth}
        \includegraphics[width=\linewidth]{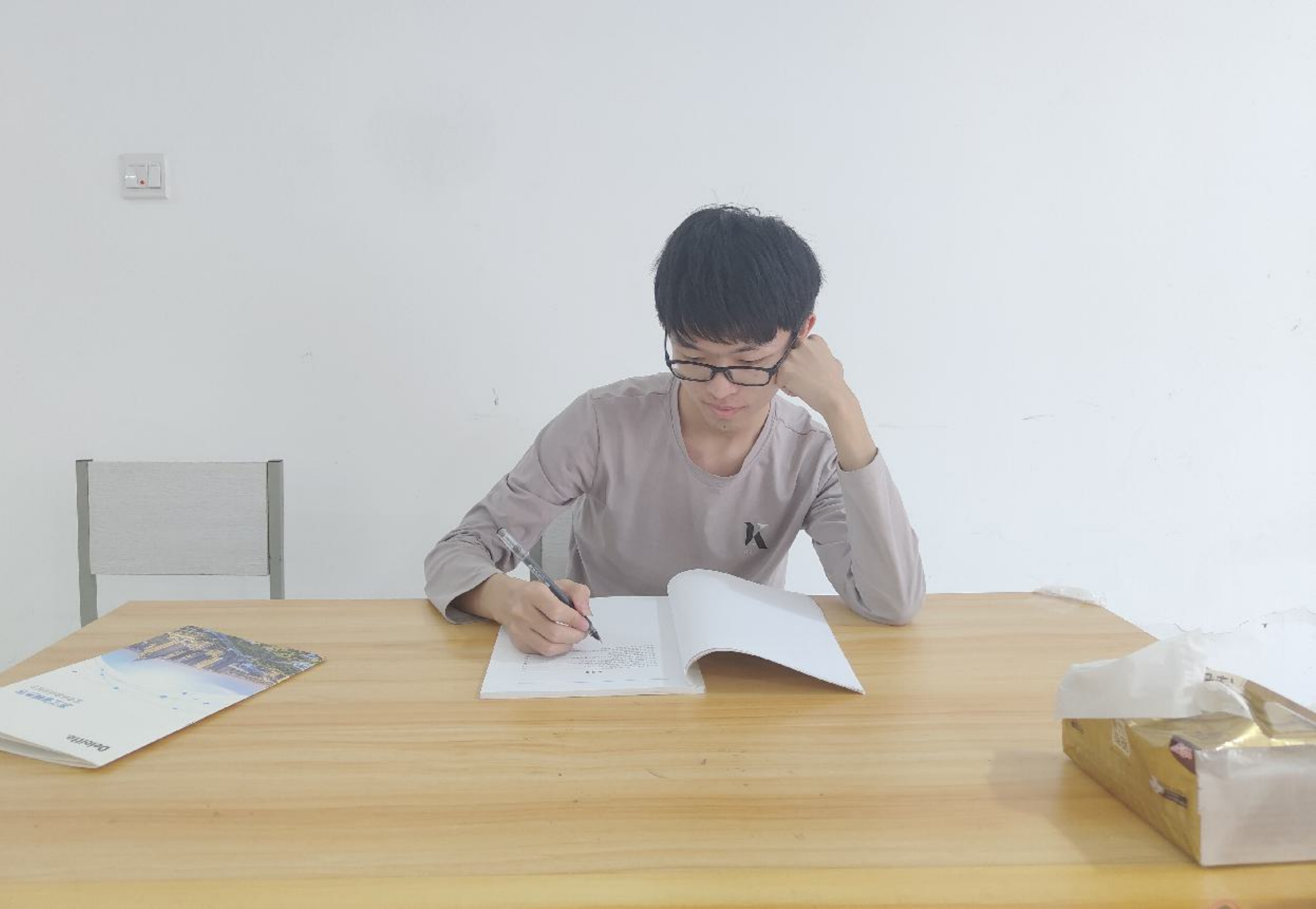}
        \caption{ChinSupport}
    \end{subfigure}
    \begin{subfigure}{0.3\columnwidth}
        \includegraphics[width=\linewidth]{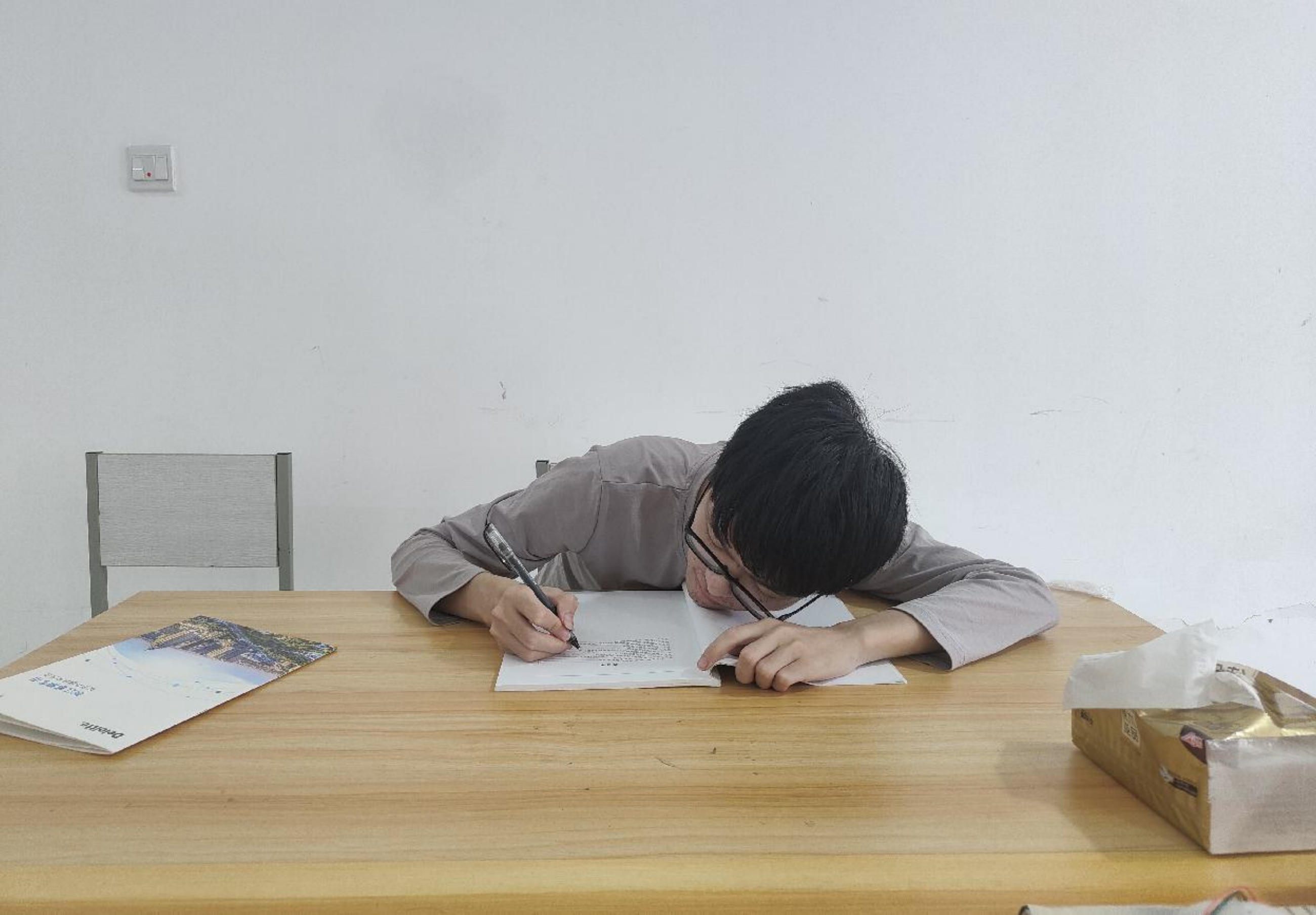}
        \caption{OnDesk}
    \end{subfigure}
    \begin{subfigure}{0.3\columnwidth}
        \includegraphics[width=\linewidth]{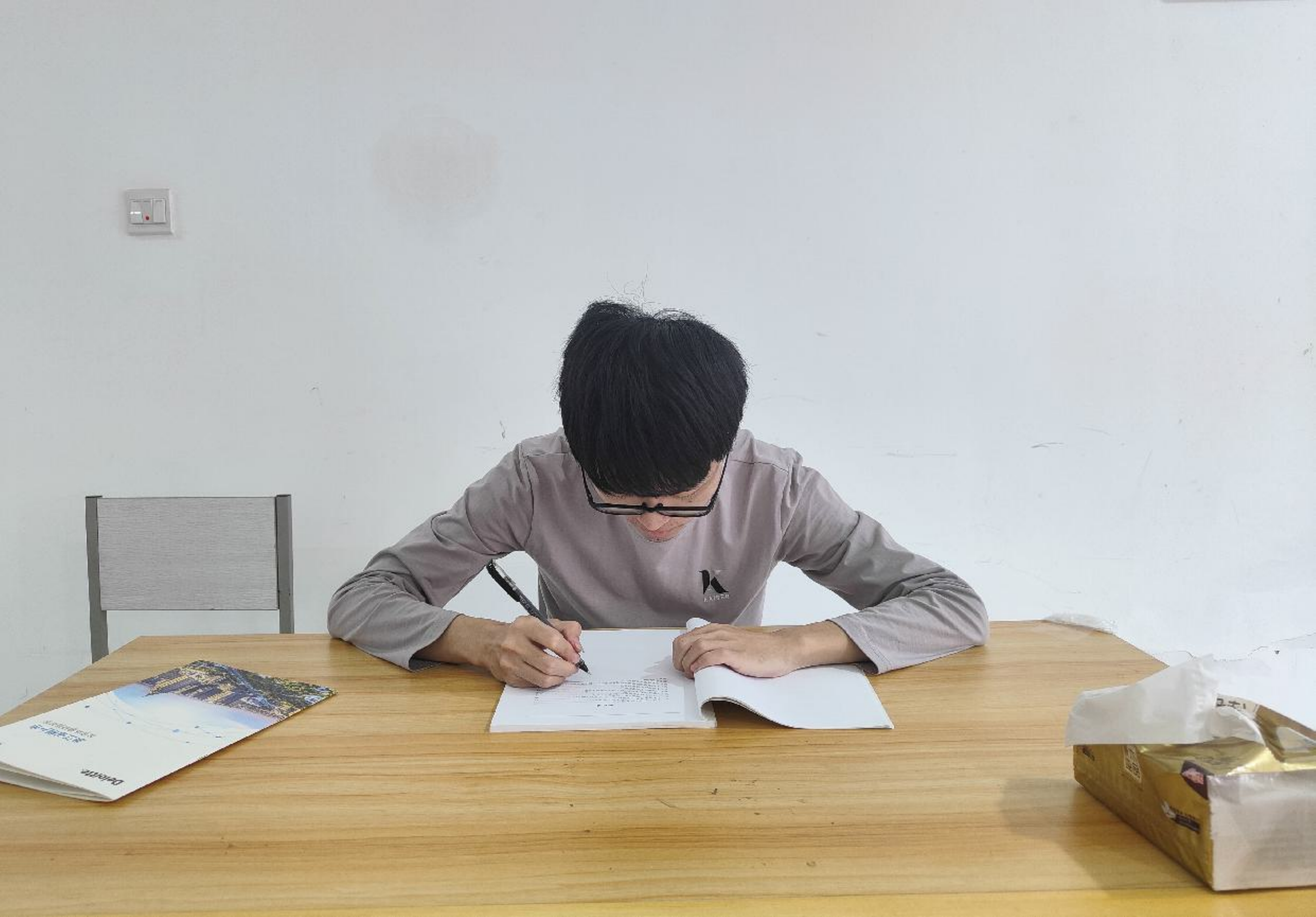}
        \caption{HeadDown}
    \end{subfigure}
    \caption{6 sitting postures}
    \label{figure:sitting posture}
\end{figure}

\begin{figure}[!h]
    \centering
    \includegraphics[width=0.6\textwidth]{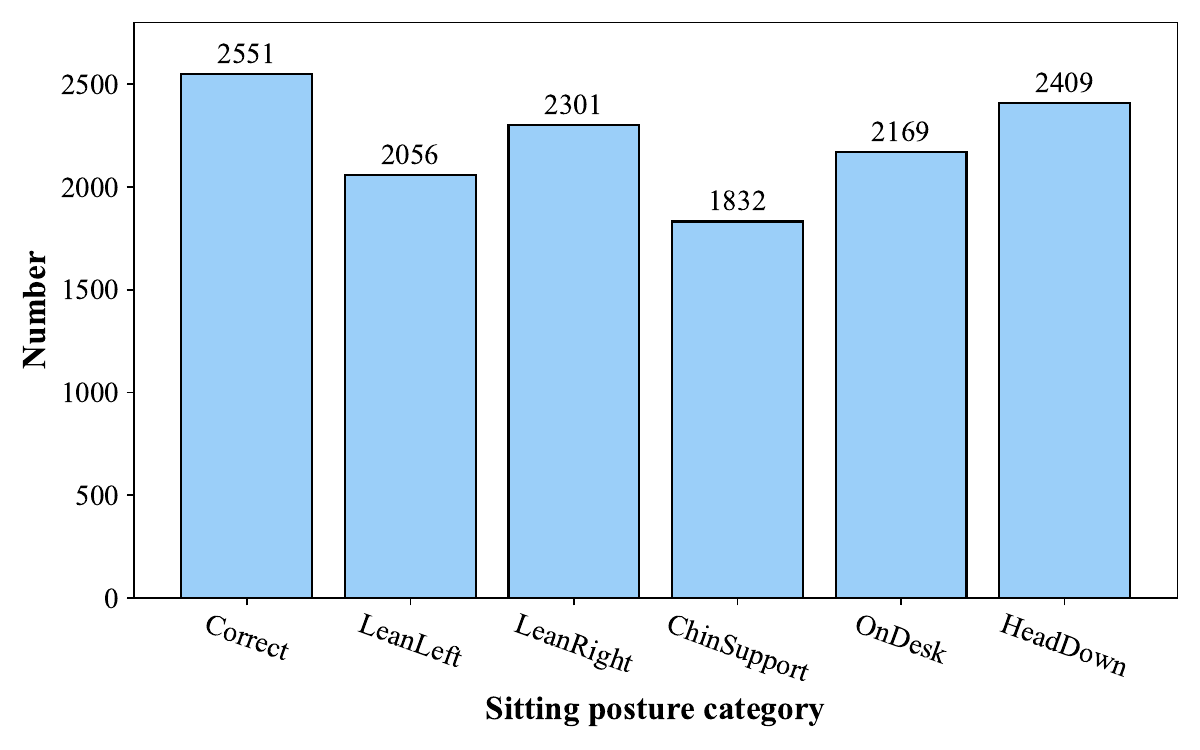}
    \caption{Statistical distribution of sitting posture categories}
    \label{figure:bar_chart}
\end{figure}

Based on ergonomics and daily office settings, six common postures were defined, comprising one correct and five incorrect postures, as follows:
\begin{enumerate}[(\arabic*)]
    \item Correct sitting posture: Maintain a naturally upright spine and keep the head at an appropriate angle, following ergonomic standards. Maintaining correct posture over time can help prevent cervical and lumbar disorders.
    \item Left-leaning / right-leaning posture: In modern office, reading, or mouse-operating scenarios, some participants tend to lean to one side. This habit can lead to scoliosis or imbalanced stress on the lumbar muscles, resulting in potential health risks.
    \item Arm-supported head posture: Some people tend to support their heads with their arms to relieve fatigue. However, this habit can cause prolonged uneven loading on the neck muscles, leading to chronic fatigue and cervical spine issues.
    \item Head-down posture: With the widespread use of mobile devices, looking down at phones has become one of the most common unhealthy postures. Prolonged head-down posture increases neck strain and the risk of cervical flexion syndrome.
    \item Desk-leaning posture: After prolonged sitting, some people tend to lean on the desk to continue working or studying. This posture can adversely affect the spine, respiratory, and digestive systems, and may even lead to reduced cerebral oxygen supply.
\end{enumerate}
\cref{figure:sitting posture} illustrates six categories of sitting postures,
and the number of images for each posture is displayed in \cref{figure:bar_chart}.

To ensure the diversity and representativeness of data,
15 participants of different genders and age groups were recruited for image collection under both single- and multi-person scenarios.
Since the lower body is often occluded by desks or other objects in real-world scenarios, the model is required to determine sitting posture solely from upper-body features.
Accordingly, 11 upper-body keypoints were precisely annotated for each image, along with the corresponding posture category and target bounding box.
Finally, a sitting posture dataset containing 5000 images was constructed and divided into training, validation, and test sets in a 7:1.5:1.5 ratio to ensure reliable training and evaluation and good generalization capability of the model.
\subsubsection{Implements details}
The experiments were conducted on Linux Ubuntu 20.04 operating system with PyTorch 1.13.0 and Python 3.8.0.
The hardware configuration featured an AMD EPYC 7742 64-core CPU and two NVIDIA RTX 3090 GPUs. Each network was trained for 300 epochs. The batch size was set to 32, the initial learning rate was fixed at 0.01, and cosine annealing was applied to decay the learning rate to $1 \times 10^{-4}$.
Each image was resized to 640 × 640 pixels, and data augmentation techniques including random scaling, random horizontal shifting, HSV color space transformation, and random flipping were applied during training to improve the model’s generalization capability. The same training configuration was used for all experimental groups.

\subsection{Performance}
First, four model scales (n, s, m, and l) were trained based on model width and depth. Model width determines the number of channels per module, while model depth corresponds to the number of modules that determine the overall model size. The results are shown in \cref{table:scales performence}.

\begin{table}[!h]
    \centering
    \caption{Performance of different model scales}
    \begin{tabular}{cccccccc}
    \hline
    Scale       & depth & width & Params(M) & GFlops & fps & mAP(\%) & Precision(\%) \\ \hline
    n &0.33       &0.25      &1.9     &4.2    &251      &61.5     &94.2           \\
    s &0.33       &0.5       &7.9     &16.4   &167      &65.6     &96.9           \\
    m &0.67       &0.75      &15.1    &39.5   &66       &68.1     &97.1           \\ 
    l &1.0       &1.0      &22.9    &75.2   &29       &71.3     &97.1           \\ \hline
    \end{tabular}
    \label{table:scales performence}
\end{table}
As the model scale increased, accuracy improved while inference speed decreased. The LSP-YOLO-n model had only 1.9M parameters, achieved the highest inference speed (251 fps) and reached a recognition accuracy of 94.2\%. 
In contrast, the l version provided the highest keypoint recognition accuracy (mAP 68.1\%) and posture classification accuracy (96.9\%) while exhibited slower inference. 
Overall, all models maintained low parameter counts and computational costs, making them suitable for deployment on various edge devices.

\begin{figure}[!h]
    \centering
    \includegraphics[width=0.8\textwidth]{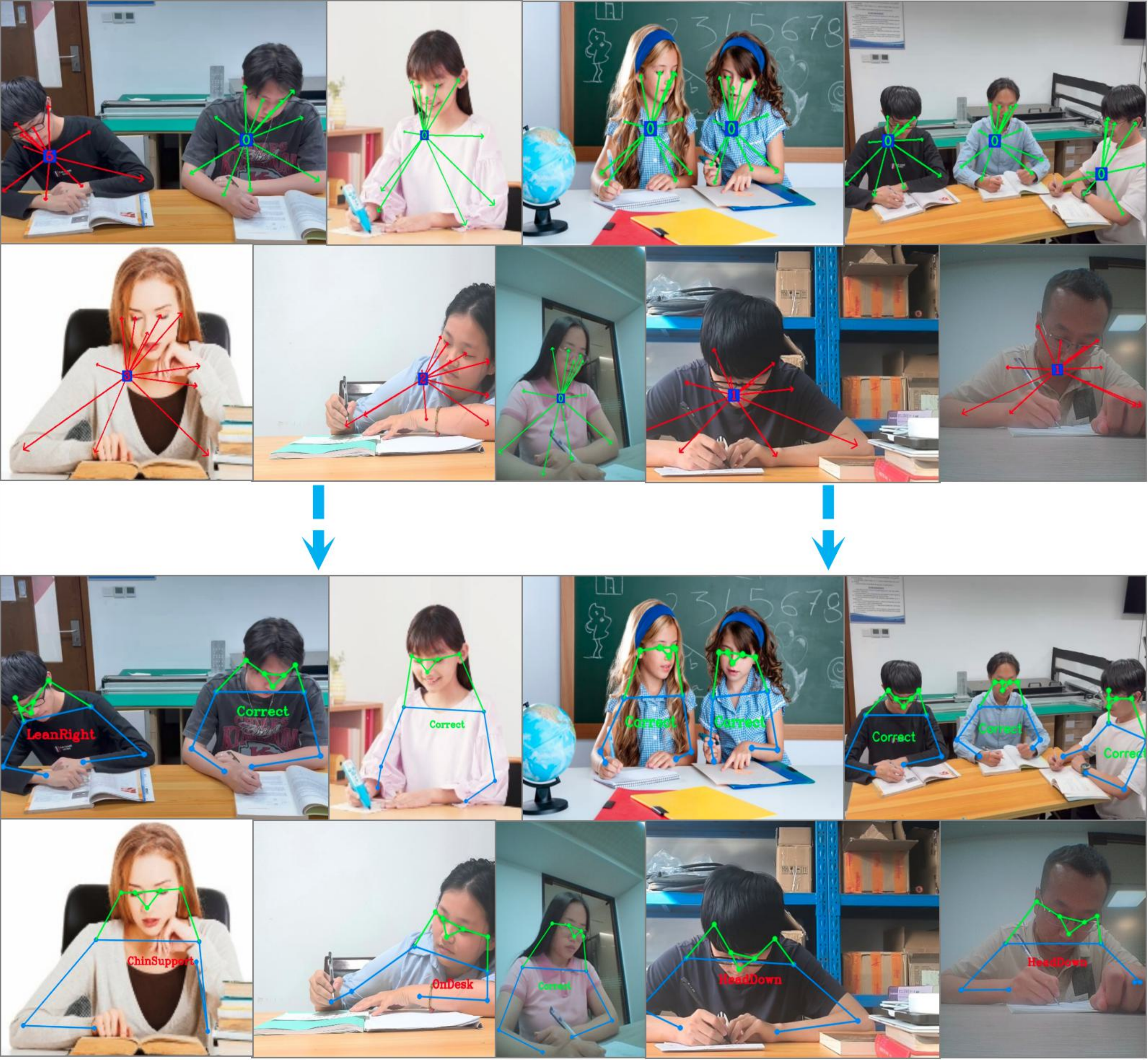}
    \caption{Visualization of inference results with LSP-YOLO-s}
    \label{figure:Test results}
\end{figure}

To intuitively illustrate the algorithm’s inference performance, inference results obtained using the LSP-YOLO-s model are presented in \cref{figure:Test results}.
\subsection{Comparative experiments}

To verify the performance advantages of the proposed algorithm, LSP-YOLO was comprehensively compared with conventional two-stage posture detection methods that rely on human keypoint information. The comparison methods included combinations of various keypoint extraction networks and independent posture classification models, including Naïve Bayes (NB), Convolutional Neural Networks (CNN), Multinomial Logistic Regression (MLR), and Support Vector Machines (SVM). The experimental results are summarized in \cref{table:performence comparison}.
\begin{table}[!h]
    \centering
    \caption{Comparison between LSP-YOLO and conventional two-stage posture detection methods. LSP-YOLO-s* removes the posture classification component of LSP-YOLO and performs only keypoint evaluation.}
    \begin{tabular}{cccccc}
    \hline
    Method           & Params(M) & GFlops & Fps & mAP(\%) & Percision(\%) \\ \hline
    LSP-YOLO-s        &7.9         &\textbf{16.4}       &\textbf{167}              &65.6     &\textbf{96.9}           \\
    LSP-YOLO-s*+SVM   &7.9         &16.4       &74               &65.9     &89.3           \\
    LSP-YOLO-s*+CNN   &8.3         &16.4       &61               &65.9     &92.5           \\
    LSP-YOLO-s*+NB    &7.9         &16.4       &81               &65.9     &79.3           \\
    LSP-YOLO-s*+MLR  &7.9         &16.4       &83               &65.9     &85.7           \\
    \makecell{YOLOv11-Pose-s\\\citep{maji2022yolo}\\+CNN} &10.9       &28.1       &51                   &66.1     &91.3           \\
    \makecell{Lightweight-Openpose\\\citep{osokin2018real}+CNN} &\textbf{4.1}       &54.1       &16                   &50.1     &82.9           \\
    \makecell{CenterNet\citep{zhou2019objects}\\+CNN}        &19.9     &202.0       &34              &59.1   &87.9       \\
    \makecell{KAPAO-s\citep{mcnally2022rethinking}\\+CNN}        &12.9     &17.1       &24              &65.1   &90.9       \\ 
    \makecell{RTMO-s\citep{lu2024rtmo}\\+CNN}        &9.7     &24.3       &53              &\textbf{68}   &92.9       \\ \hline
\end{tabular}
    \label{table:performence comparison}
\end{table}

The results showed that since the proposed method no longer relied on an additional classification model, the overall model size was significantly reduced and the inference speed was much higher than traditional two-stage approaches.
Although LSP-YOLO did not achieve the highest mAP, it outperformed other methods in posture classification accuracy. This may be due to the end-to-end training paradigm, where keypoint extraction and classification were jointly optimized in one network, allowing the classifying part to exploit richer features than in two-stage methods, which rely only on compressed keypoint coordinates.

On the other hand, intermediate supervision slightly reduced keypoint accuracy but significantly improved posture classification performance and inference efficiency, making this trade-off valuable for embedded deployment scenarios.

\subsection{Module ablation experiments}
To evaluate the impact of the Light-C3k2 module on network resource usage and detection performance, ablation experiments were conducted. Changes in key metrics, including model size, inference memory usage, inference speed and detection accuracy  were compared before and after introducing PConv and SimAM. The results are shown in the \cref{table:Light-C3k2 ablation experience}.
\begin{table}[!h]
    \centering
    \caption{Results of ablation experiments}
    \begin{tabular}{ccccccc}
    \hline
    PConv & SimAM & Size(M) & Gflops &Fps & mAP(\%) & Precision(\%)         \\ \hline
    \xmark     & \xmark     &9.0   &19.4    &103     &66.2     &96.3           \\
    \cmark     & \xmark     &7.9   &16.4    &\textbf{176}     &63.7     &94.6           \\
    \xmark     & \cmark     &7.9   &19.4    &101     &\textbf{66.9}     &96.9           \\
    \cmark     & \cmark     &\textbf{7.9}   &\textbf{16.4}    &167     &65.6     &\textbf{96.9}           \\ \hline
    \end{tabular}
    \label{table:Light-C3k2 ablation experience}
\end{table}
Compared with the baseline, introducing only PConv improved inference speed by 73 fps but reduced mAP and precision by 2.9\% and 1.7\%, respectively.
These results suggest that although PConv effectively improved inference efficiency, it caused some loss of detection accuracy.
Meanwhile, introducing only SimAM increased precision and mAP by 0.6\% and 0.7\%, respectively, while inference speed remained unchanged. These results suggest that SimAM enhanced feature representation without extra computation.
When PConv and SimAM were introduced together, the inference speed improved by 64 fps compared with the baseline, and precision increased by 0.6\%. Although the mAP decreased by 0.6\% compared with the baseline, precision and mAP increased by 2.3\% each relative to using only PConv. These results suggest that SimAM effectively compensated for the accuracy loss caused by PConv, enabling the model to maintain detection performance comparable to the baseline while sustaining high inference speed.

\begin{figure}[!h]
    \centering
    \includegraphics[width=\textwidth]{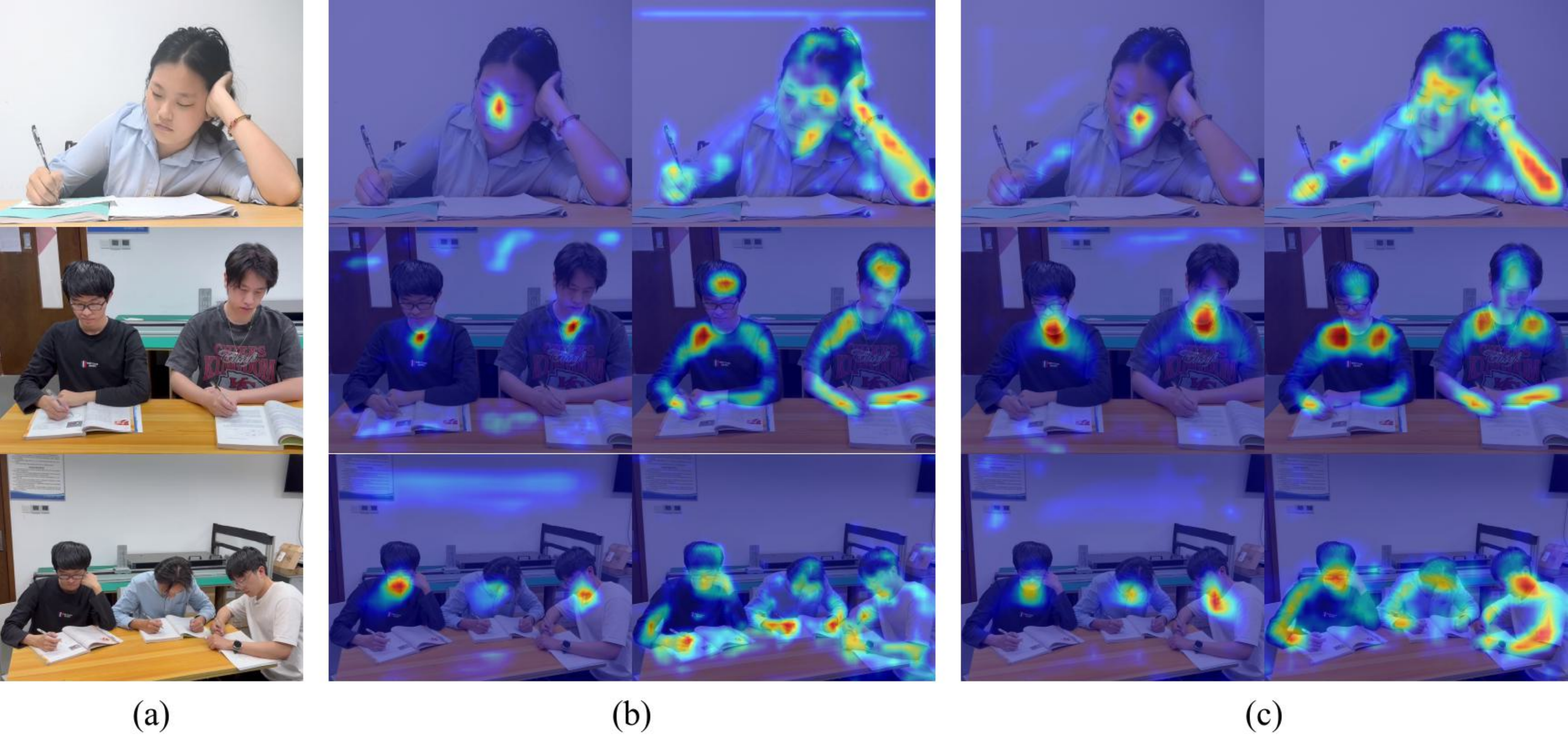}
    \caption{Grad-CAM visualizations of models with and without SimAM. (a) 
Input images; (b)Model with SimAM: activations are more focused on upper-body keypoints; (c) Model without SimAM: activations are dispersed and less relevant to posture features.}
    \label{figure:gradcam}
\end{figure}

Furthermore, 
to visually illustrate the optimization effect of the SimAM module within the network architecture, 
the gradient-weighted class activation mapping heatmaps were visualized, 
as shown in \cref{figure:gradcam}. 
The high-activation regions detected by all tested models were mainly distributed around the keypoints of the upper body. 
Compared with the model without SimAM, 
the SimAM-enhanced model exhibited more precise and concentrated attention on the target areas. 
These results demonstrate that introducing this parameter-free attention mechanism can effectively enhance the model’s feature extraction capability, 
enabling more efficient extraction and utilization of semantic information.

\subsection{Experiments on loss weights}
To further verify the role of intermediate supervision in the sitting posture recognition network and the influence of its weight settings on the overall performance, this section conducted a systematic scanning experiment on the sitting posture recognition loss weight $\alpha$ and the posture estimation loss weight $\beta$.
Specifically, the confidence weight $\gamma$ was fixed at 1, and under identical training conditions, different values of $\alpha$ and $\beta$ were tested to evaluate their impact on the overall sitting posture recognition accuracy.
The experimental results were presented in \cref{table:weight experienments}.
\begin{table}[!h]
    \centering
    \setlength{\tabcolsep}{12pt}
    \caption{Results of weight experiments}
    \begin{tabular}{cccc}
    \hline
    $\alpha$ & $\beta$ & mAP(\%) & Precision(\%) \\ \hline
    2                     & 8                    &64.5     &91.8          \\
    4                     & 8                    &64.2     &92.5          \\
    6                     & 8                    &64.1     &92.7          \\
    2                     & 10                   &\textbf{65.7} &93.5              \\
    \textbf{4}            & \textbf{10}          &65.6     &\textbf{96.9}          \\
    6                     & 10                   &65.6     &95.0          \\
    2                     & 12                   &65.4     &93.5          \\
    4                     & 12                   &65.3     &93.7          \\
    6                     & 12                   &65.6     &92.7               \\ \hline
    \end{tabular}
    \label{table:weight experienments}
\end{table}

The experimental results showed that the weight configuration affected the degree of task coupling during training. When $\beta$ increased to 12, both keypoint detection accuracy and sitting posture classification accuracy improved, indicating that strengthening intermediate supervision in the keypoint branch facilitated overall feature learning and integration within the network.
As $\beta$ continued to increase, however, the performance gain gradually saturated, suggesting that excessive intermediate supervision imposed constraints that limited the optimization of the classification branch.
Proper adjustment of the classification loss weight enhanced the accuracy of sitting posture classification.
The experimental results showed that when $\alpha$ : $\beta$ = 4 : 10, the network achieved the best overall recognition accuracy.
This finding validated the effectiveness of the intermediate supervision mechanism in facilitating the coupling between keypoint and sitting posture features.

\subsection{Embedded deployment experiments}
To evaluate the deployment potential of LSP-YOLO on embedded edge devices and assess its performance in real-world scenarios, a resource-constrained hardware platform was built. Comparative experiments were then conducted between LSP-YOLO and the two-stage approach based on LSP-YOLO*+CNN.

\begin{figure}[!h]
    \centering
    \includegraphics[width=\textwidth]{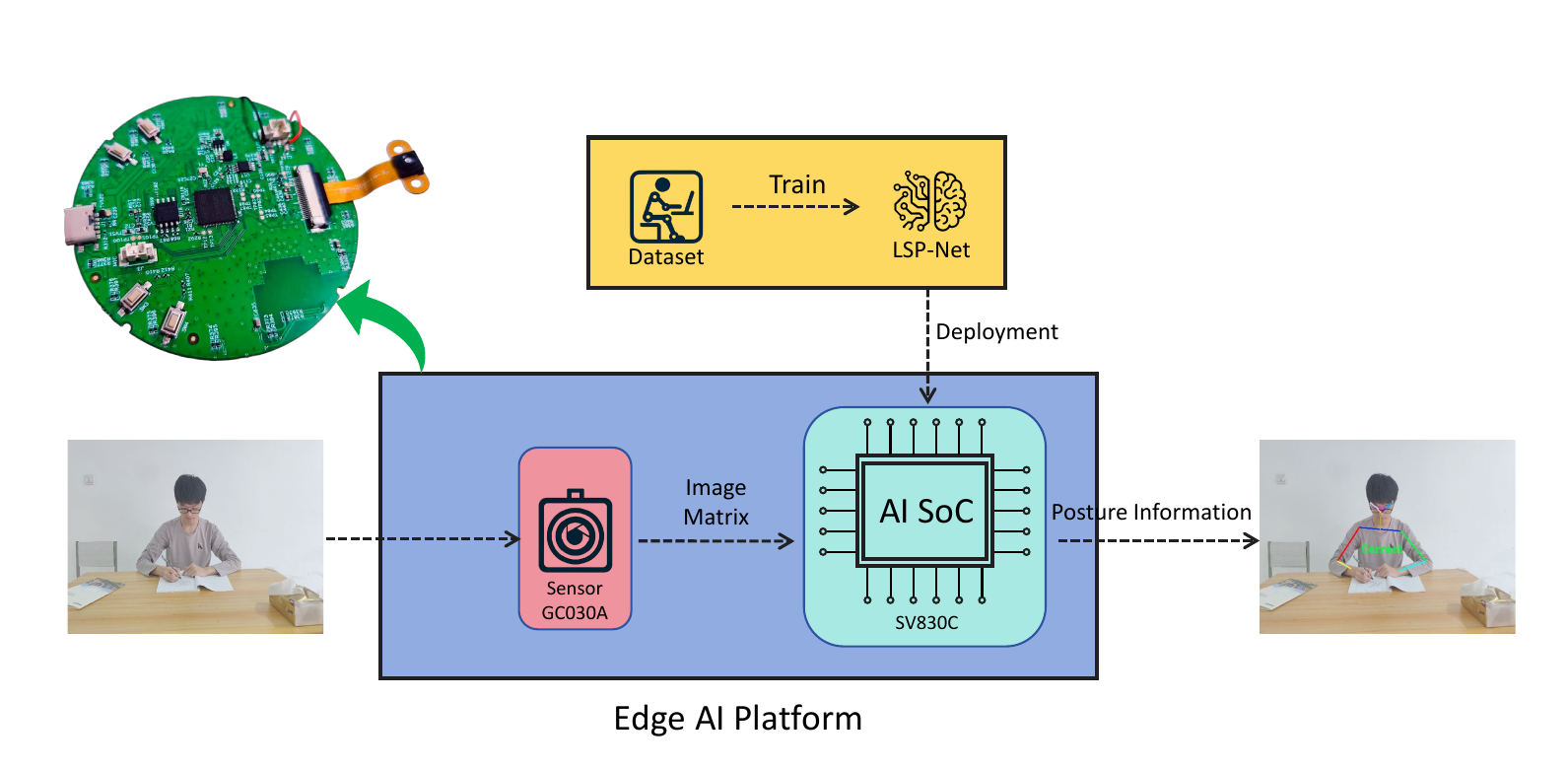}
    \caption{Architecture of the embedded testing platform for posture recognition}
    \label{figure:deploy_system}
\end{figure}

Specifically, the SV830C chip and GC030A image sensor were selected to build an embedded edge computing platform for model quantization deployment and inference testing. The SV830C chip featured 16 MB Flash, 64 MB RAM, and 0.5-TOPS AI computing power.
The GC030A provided maximum resolution of 640×480 and frame rate of 30 fps.
The overall structure of the testing system is shown in \cref{figure:deploy_system}.
We first trained the model on PC, then used the quantization tool provided for the SV830C chip to quantize the model and deploy it onto the chip. 
The model performed forward inference on images captured by the sensor.

\begin{figure}[!h]
    \centering
    \includegraphics[width=0.89\textwidth]{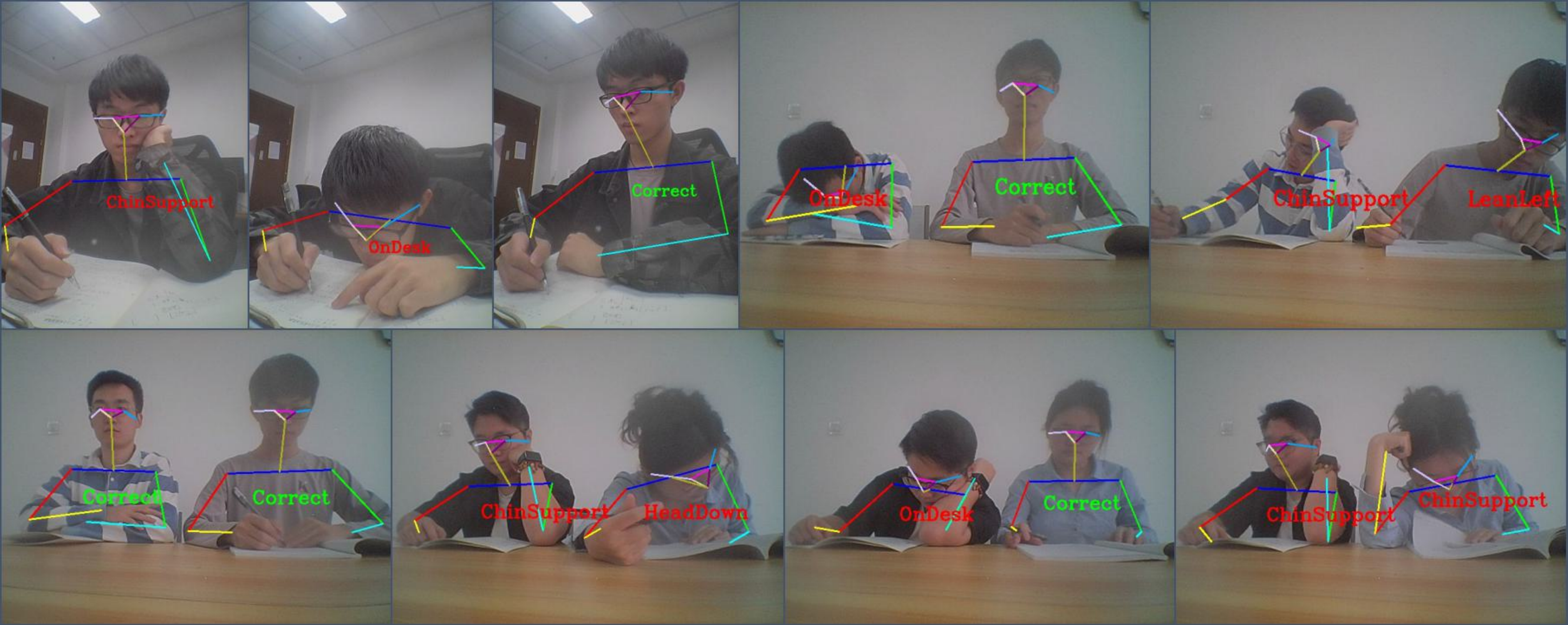}
    \caption{Visualization of sitting posture recognition results on embedded platform}
    \label{figure:embedded_results}
\end{figure}

To obtain stable and reliable runtime performance metrics, 1000 consecutive inference runs were performed on the edge platform, and the chip’s resource utilization including memory usage, inference latency and power consumption was recorded. The results are summarized in \cref{table:embedded_performance}. Additionally, some inference results were encoded and visualized, as shown in \cref{figure:embedded_results}.
\begin{table}[!h]
    \centering
    \caption{Performance evaluation results on embedded edge device. LSP-YOLO-s* removes the posture classification component of LSP-YOLO and performs only keypoint evaluation.}
    \begin{tabular}{ccccccc}
    \hline
Method          & Precision(\%) & \begin{tabular}[c]{@{}c@{}}Preprocessing \\ latency(ms)\end{tabular} & \begin{tabular}[c]{@{}c@{}}Inference \\ latency(ms)\end{tabular} & \begin{tabular}[c]{@{}c@{}}Memory \\ usage(M)\end{tabular} & \begin{tabular}[c]{@{}c@{}}Storage \\ usage(M)\end{tabular} \\ \hline
LSP-YOLO-n      & 91.7      & 115  & 255  & 22  & 2.2 \\
\begin{tabular}[c]{@{}c@{}}LSP-YOLO-n*\\ +CNN\end{tabular} & 81.3      & 117  & 637  & 26  & 2.4 \\ \hline

\end{tabular}
\label{table:embedded_performance}
\end{table}

\begin{figure}[!h]
    \centering
    \includegraphics[width=0.7\textwidth]{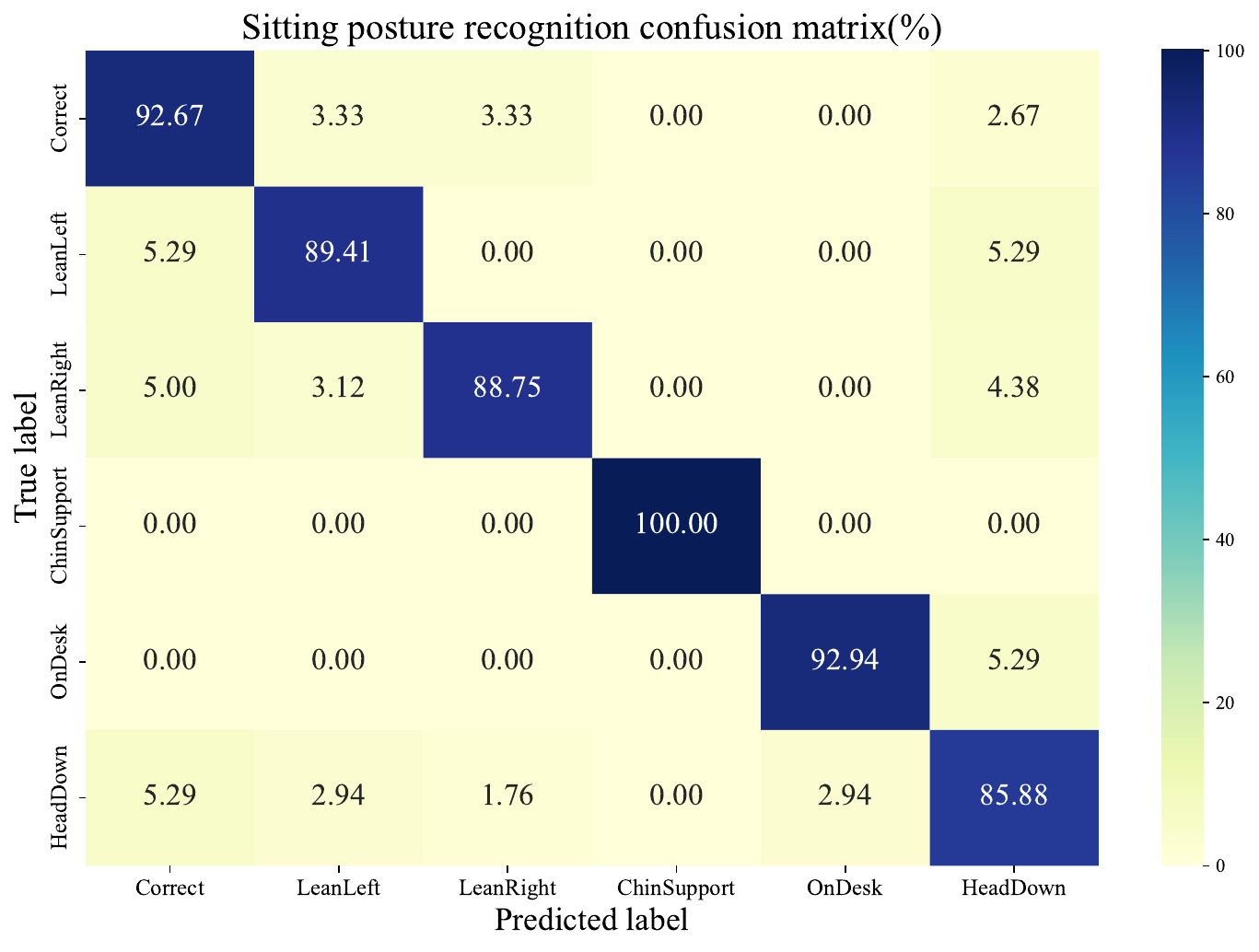}
    \caption{Classification confusion matrix of quantized model}
    \label{figure:matrix}
\end{figure}

Compared with its performance on the PC platform, the two-stage method showed an 10.4\% accuracy drop and a significant increase in inference latency after deployment on the edge device. This was mainly because quantization introduced distortions to the keypoint outputs of the first stage, which accumulated through the second-stage classification model and led to degraded overall accuracy.
Moreover, frequent storage and retrieval of intermediate data in the two-stage method caused repeated memory access, further increasing overall inference latency.
In contrast, as a single-stage integrated model, 
LSP-YOLO experienced only a 5.2\% decrease in accuracy after quantization. Moreover, because it required no intermediate data processing, 
it achieved a substantially higher inference speed than two-stage methods. 
These results demonstrated the efficiency and superiority of the proposed approach in edge computing scenarios.
We evaluated the test results of the quantized LSP-YOLO 
and the false positive samples were illlustrated in the confusion matrix, as shown in \cref{figure:matrix}.
\section{Conclusions}
This paper proposed a lightweight posture recognition neural network, LSP-YOLO, designed for deployment on edge devices. It achieved fast and accurate posture recognition with minimal storage and computational resources, helping users effectively avoid health risks caused by improper sitting postures.

The core innovation of this study lied in the integration of two traditionally independent stages in keypoints extraction and posture classification into a single, trainable end-to-end network structure for the first time.
By introducing point convolution in the network head, the model achieved seamless fusion of keypoint information and posture classification, substantially reducing computational and storage overhead. In addition, for resource-constrained embedded deployment scenarios, a lightweight feature extraction module, Light-C3k2, was designed. 
Incorporating the concept of PConv and SimAM,  
it effectively reduced the memory access frequency during inference, further improving 
the inference speed and hardware adaptability.

Experimental results showed that LSP-YOLO outperformed traditional two-stage methods by achieving smaller model size, lower resource consumption and higher inference efficiency while maintaining high accuracy. Furthermore, the deployment tests on the embedded edge platform under limited resources demonstrated that LSP-YOLO delivered stable and accurate real-time posture recognition, which confirmed its feasibility and efficiency in real-world applications.

Overall, this study proposed an efficient and novel solution for sitting posture recognition, effectively addressing the issues of large model size and error accumulation inherent in traditional two-stage methods. In addition, the proposed approach provided a useful reference for other keypoint-based downstream tasks (e.g., violence detection and fall detection) through an end-to-end lightweight modeling framework.
\bibliographystyle{elsarticle-harv}  
\bibliography{refs}
\end{document}